\documentclass[10pt,twocolumn,letterpaper]{article}

\usepackage{wacv}
\usepackage{times}
\usepackage{epsfig}
\usepackage{graphicx}
\usepackage{amsmath}
\usepackage{amssymb}

\usepackage{bbm}
\usepackage[margin=0pt,font=small,labelfont=small,justification=justified,labelsep=period]{caption}
\usepackage{adjustbox}
\usepackage{multirow}
\usepackage{booktabs}
\usepackage{xspace}
\usepackage{color}
\usepackage{multirow}
\usepackage{graphics}
\usepackage{adjustbox}
\usepackage{subfigure}
\usepackage{amsmath}
\usepackage{paralist} % compactitem, compactenum
\usepackage{enumitem} % adjust enum margin
\usepackage{booktabs} % Publication quality tables
\usepackage{array}
\usepackage{grffile}
\usepackage{makecell}
\usepackage{diagbox}

\graphicspath{{figures/}}
\usepackage[pagebackref=true,breaklinks=true,letterpaper=true,colorlinks,bookmarks=false,citecolor=blue,linkcolor=blue]{hyperref}

\newcommand{\Paragraph}[1]{{\vspace{-2mm}\flushleft\textbf{#1}}} % avoid indent for

\long\def\ignorethis#1{}

\usepackage{tabularx}

\definecolor{B1}{RGB}{237,219,201}
\definecolor{MyBlueb}{RGB}{95,169,236}
\definecolor{MyOrange}{RGB}{255,177,98}
\definecolor{MyReda}{RGB}{193,39,45}
\definecolor{gray}{rgb}{0.5,0.5,0.5}
\definecolor{MyBlue}{rgb}{0,0,1.0}
\definecolor{MyYellow}{rgb}{0.9,0.9,0}
\definecolor{MyRed}{rgb}{0.8,0.2,0}
\definecolor{MyGreen}{rgb}{0,0.5,0.0}
\definecolor{MyGray}{rgb}{0.4,0.4,0.4}

\def\red#1{\textcolor{MyRed}{#1}}
\def\blue#1{\textcolor{MyBlue}{#1}}

\def\first#1{\red{\textbf{#1}}}
\def\second#1{\blue{\underline{#1}}}

\newlength\paramargin
\newlength\figmargin
\newlength\secmargin

\newcolumntype{L}[1]{>{\raggedright\let\newline\\\arraybackslash\hspace{0pt}}m{#1}}
\newcolumntype{C}[1]{>{\centering\let\newline\\\arraybackslash\hspace{0pt}}m{#1}}
\newcolumntype{R}[1]{>{\raggedleft\let\newline\\\arraybackslash\hspace{0pt}}m{#1}}

\def\eg{e.g.,~}

\def\etal{et~al.\xspace}

\newcommand{\figref}[1]{Figure~\ref{#1}}
\newcommand{\tabref}[1]{Table~\ref{#1}}
\newcommand{\eqnref}[1]{Equation~\eqref{#1}}

\makeatletter

\setbox0\hbox{$\xdef\scriptratio{\strip@pt\dimexpr
		\numexpr(\sf@size*65536)/\f@size sp}$}

\newcommand{\scriptveryshortarrow}[1][5pt]{{%
		\hbox{\rule[\scriptratio\dimexpr\fontdimen22\textfont2-.2pt\relax]
			{\scriptratio\dimexpr#1\relax}{\scriptratio\dimexpr.4pt\relax}}%
		\mkern-5mu\hbox{\let\f@size\sf@size\usefont{U}{lasy}{m}{n}\symbol{41}}}}

\makeatother

\def\wacvPaperID{253} % Enter the WACV Paper ID here

\wacvfinalcopy % *** Uncomment this line for the final submission

\ifwacvfinal
\def\assignedStartPage{1} % *** Enter the assigned starting page number (instead of 9876)
\fi

\ifwacvfinal
\usepackage[breaklinks=true,bookmarks=false]{hyperref}
\else
\usepackage[pagebackref=true,breaklinks=true,colorlinks,bookmarks=false]{hyperref}
\fi

\ifwacvfinal
\else
\fi

\begin{document}

%%%%%%%%% TITLE
%\title{Prediction-assistant Video Frame Super-Resolution}

\title{Prediction-assistant Frame Super-Resolution for Video Streaming}

% \and

\author{Wang Shen$^1$ 
\hspace{3pt}
Wenbo Bao$^1$
\hspace{3pt}
Guangtao Zhai$^1$ $^\ast$
\hspace{3pt}
Charlie L Wang$^2$
\hspace{3pt} 
Jerry W Hu$^2$
\hspace{3pt} 
Zhiyong Gao$^1$
\\
$^1$ Institute of Image Communication and Network Engineering, \\ Shanghai Jiao Tong University \\
$^2$ Intel
\vspace{-6mm}
}

\maketitle
%\thispagestyle{empty}

%%%%%%%%% ABSTRACT
\begin{abstract}
    Video frame transmission delay is critical in real-time applications such as online video gaming, live show, etc.
    The receiving deadline of a new frame must catch up with the frame rendering time.
    Otherwise, the system will buffer a while, and the user will encounter a frozen screen, resulting in unsatisfactory user experiences.
    An effective approach is to transmit frames in lower-quality under poor bandwidth conditions, such as using scalable video coding.
    In this paper, we propose to enhance video quality using lossy frames in two situations.
    First, when current frames are too late to receive before rendering deadline (i.e., lost), we propose to use previously received high-resolution images to predict the future frames.
    Second, when the quality of the currently received frames is low~(i.e., lossy), we propose to use previously received high-resolution frames to enhance the low-quality current ones.
    For the first case, we propose a small yet effective video frame prediction network.
    For the second case, we improve the video prediction network to a video enhancement network to associate current frames as well as previous frames to restore high-quality images.
    Extensive experimental results demonstrate that our method performs favorably against state-of-the-art algorithms in the lossy video streaming environment.
    We will publish the source code and the pre-trained models upon acceptance.
    %
    % {\let\thefootnote\relax\footnote{This work was performed when the first author was an intern at Intel. \\ $^\ast$ Corresponding author}}
\end{abstract}

%%%%%%%%% BODY TEXT
\section{Introduction}
Video streaming is a key technology in broadcasting industries~\cite{index2016white}, such as online video gaming, live show, video on demand, etc.
Due to the variability of the video transmission channel, it is difficult to ensure the user-perceived Quality of Experience (QoE)~\cite{balachandran2013developing, krishnan2013video, dobrian2011understanding}.
Two main factors affect video QoE.
One is the quality of received frames, while the other is the transmission delay.
Since the randomness of time-varying channel fluctuations degrades the transmission performance, the streaming system will encounter a buffering interruption.
Then users may encounter a frozen screen, resulting in a dramatic decrease in QoE.
A plausible way is to transmit frames in lower-quality when encountering poor bandwidths, such as using scalable video coding~\cite{schwarz2007overview}.
In this paper, we consider one of the image quality degradations, namely spatial resolution degradation.
A low-resolution image with fewer pixels saves more transmission bits, thus reducing the possibility of buffering interruption.
However, transmitting low-resolution images with fewer pixels will also reduce the quality of received frames.
Therefore, high image quality and low transmission delay are contradictory to each other in video streaming.
Several works~\cite{yin2015control, tian2012towards, mao2017neural} propose video bitrate control algorithms to balance the trade-off between quality and delay.
Those methods require meticulous modeling.
Thus, those approaches are not adaptive to practical scenarios.

\begin{figure*}[t]
	\footnotesize
    % \scriptsize

	\centering
% 	\renewcommand{\tabcolsep}{1.0pt} % adjust horizontal space
% 	\renewcommand{\arraystretch}{1.0} % adjust vertical space
% 	\begin{tabular}{c}
	\includegraphics[width= 1.0\textwidth]{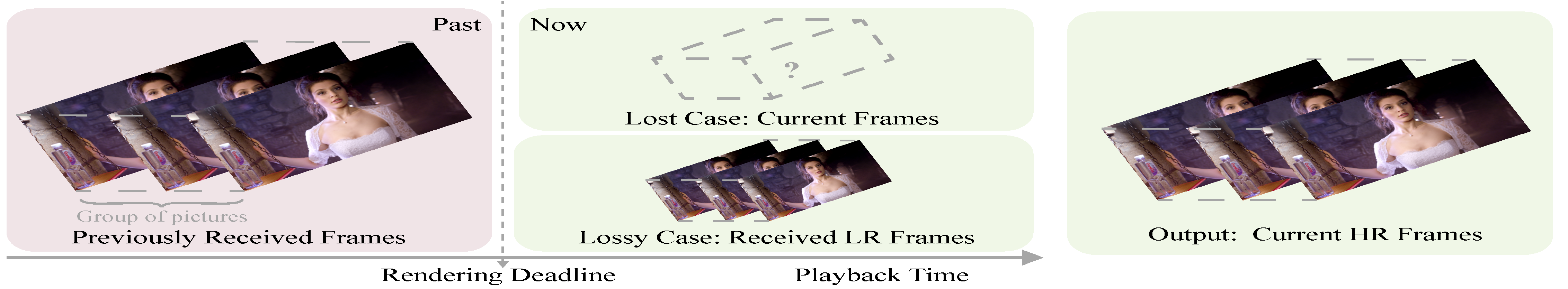}
	
% \end{tabular}
%  	\vspace{-5pt}
	\caption{
		\textbf{Illustration of lossy video streaming.} 
		We propose to restore high-resolution (HR) frames at the current time under two streaming patterns.
		First, when current frames are too late to receive before rendering deadline, we proposed to use previously received high-resolution images to predict the future frames.
        Second, when the resolution of the current frames is low (i.e., LR), we propose to use previous high-resolution frames to enhance the current low-quality frames.
	}
%  	\vspace{-10pt}
	\label{fig:arch} 
\end{figure*}

In contrast to the methods that rely on bitrate control~\cite{yin2015control, tian2012towards, mao2017neural}, we propose to improve video QoEs through low-level image processing.
We consider enhancing the quality of the received frames under two streaming patterns.
First, when current frames are too late to receive before rendering deadline (i.e., lost), we propose to use previously received high-resolution images to predict future frames.
Second, when the quality of the currently received frames is low~(i.e., lossy), but the rendering deadline is approaching, we propose to use previously received high-resolution frames to enhance the low-quality current frames.

For the first case, similar to existing video prediction algorithms~\cite{liu2017video, gao2019disentangling,liu2018future, hao2018controllable}, we propose a small, yet effective video prediction network. 
The common approach for the convolutional neural network (CNN) based video prediction network is to first predict the optical flow between the current reference frame and the future frame.
Then, warping the reference frame to the future frame based on the flow.
Finally, the occlusion areas are filled using a CNN-based synthesis network.
For this approach, the accuracy of the future optical flow prediction is significant in this framework.
Different from those works using a CNN (\eg U-Net) to directly predict the optical flow, we design a new flow predictor.
The new flow predictor integrates previously received high-resolution frames as well as wrapped flow and propagated flow to synthesize the final future flow.

\begin{figure*}[t]
	\footnotesize
    % \scriptsize

	\centering
% 	\renewcommand{\tabcolsep}{1.0pt} % adjust horizontal space
% 	\renewcommand{\arraystretch}{1.0} % adjust vertical space
% 	\begin{tabular}{c}
	\includegraphics[width= 1.0\textwidth]{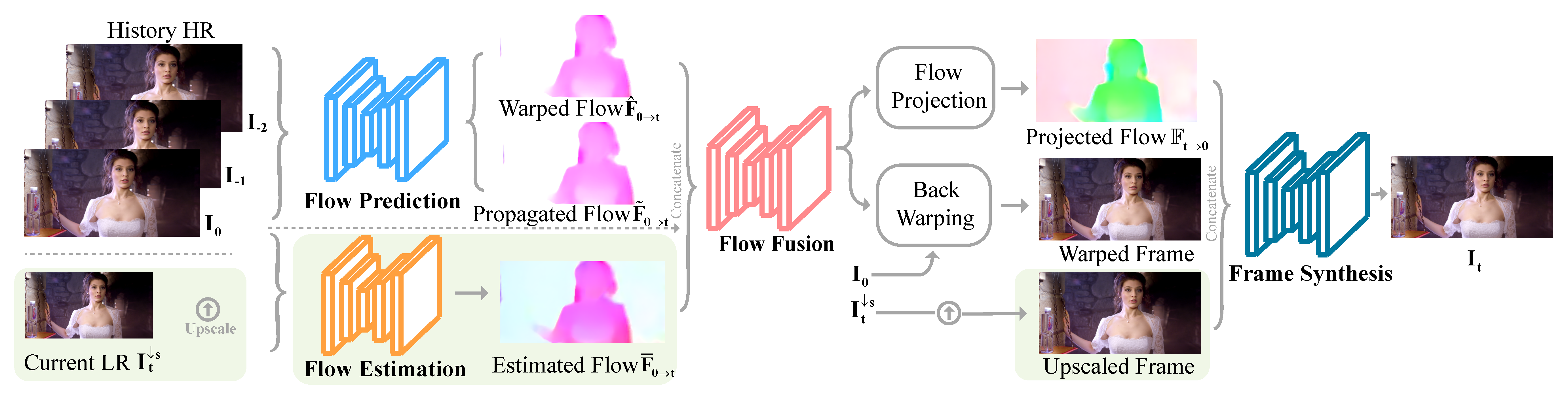}
	
% \end{tabular}
%  	\vspace{-5pt}
	\caption{
		\textbf{Architecture of the proposed prediction-assistant video frame super-resolution model.} 
		Given three previously received HR frames $\mathbf{I}_{-2}, \mathbf{I}_{-1},\mathbf{I}_0$, and one current LR frame $\mathbf{I}_t^{\downarrow s}$, where $s$ is the scale factor.
		We first use the three HR frames to predict two optical flows, including the warped flow $\mathbf{\hat{F}}_{0 \rightarrow t}$, and the propagated flow $\mathbf{\tilde{F}}_{0 \rightarrow t}$.
		We also estimate a flow $\mathbf{\bar{F}}_{0 \rightarrow t}$ using HR frame $\mathbf{I}_0$ and the up-scaled current LR frame $\texttt{bic}(\mathbf{I}_t^{\downarrow s})$.
		Then, the three flows are refined by a flow fusion network.
		We then project the fused flow and back-warp the frame $\mathbf{I}_0$ based on the projected flow $\mathbb{F}_{t \rightarrow 0}$.
		Finally, we apply a frame synthesis network to generate the output frame.
		When the current LR frame $\mathbf{I}_t^{\downarrow s}$ is not received before rendering deadline, the proposed model is the same except those green modules.
	}
%  	\vspace{-10pt}
	\label{fig:framework} 
\end{figure*}

For the second case, we obtain previously received high-resolution frames and current low-resolution frames.
Our goal is to synthesize a group of high-resolution current frames.
To this end, the common practice is to perform video super-resolution~\cite{wang2019edvr,Lim_2017_CVPR_Workshops,xue2019video,jo2018deep,bao2018memc}.
State-of-the-art video super-resolution methods typically take multiple low-resolution frames as input.
The outputs of those approaches are often blurry or too smooth, i.e., losing too much texture details.
Recently, reference-based image super-resolution algorithms are widely used in multi-camera systems~\cite{cheng2020dual,zheng2018crossnet,paliwal2020deep,yang2020learning}, which transfers high-resolution textures from the given reference image to produce visually pleasing results.
However, if the resolution of the received frame is small enough, those algorithms can not estimate an accurate future optical flow, as described in~\figref{fig:acc_scale}. 
To this end, we propose a new flow predictor that integrates the estimated flow as well as another two kinds of predicted flow, called propagated flow and wrapped flow (as described in~\figref{fig:framework}).
Thus, even if the estimated flow is not accurate, the other two flows help the flow fusion network produce an accurate optical flow.

The proposed \textbf{P}rediction-\textbf{ASS}istant Network (PASS-Net) restores high-quality frames in the lossy streaming environment.
Extensive experiments on multiple benchmarks, including the Middlebury~\cite{baker2011database}, UCF101~\cite{soomro2012ucf101}, Vimeo90K~\cite{xue2019video}, Vid4~\cite{liu2013bayesian}, REDS~\cite{Nah_2019_CVPR_Workshops_SR}, HD~\cite{bao2018memc}, demonstrate that the proposed model performs favorably against state-of-the-art methods.

Our main contributions are summarized as follows:
    \begin{compactitem}
         \item 
        	We propose a prediction-assistant video frame super-resolution method to enhance the quality of video streaming.
            The method explicitly exploits previous frames to help predict or restore current frames, which contributes to accurate optical flow estimation and provides rich texture details.
        \item
            We propose to integrate the estimated flow, propagated flow, and predicted flow for accurate future flow estimation. 
            This mechanism contributes to frame synthesis especially when the current flow is lost or the resolution of the current frame is small.
        \item 
        	We demonstrate that the proposed model is more effective,
            efficient, and compact than the state-of-the-art
            approaches.
    \end{compactitem}

\section{Related Work}

    In this section, we review previous works of learning-based video super-resolution and video prediction.

    \Paragraph{Video Super-Resolution.}
    Super-resolution can be formulated as a dense image regression problem, which learns a derivable mapping function between low-resolution and high-resolution images.
    Super-resolution algorithms can be divided into two main categories, image super-resolution, and video super-resolution.

    Image super-resolution algorithms upscale the resolution of a single image.
    Dong~\textit{\etal}~\cite{dong2015image} first propose SRCNN that composes of three-layer CNNs to learn the pixel mapping function.
    Later, Dong~\textit{\etal}~\cite{dong2016accelerating} redesign SRCNN to speed up the computation.
    Kim~\textit{\etal} propose VDSR~\cite{kim2016accurate} and DRCN~\cite{kim2016deeply} using residual connections.
    The residual blocks~\cite{he2016deep} are used as common components in SRResNet~\cite{ledig2017photo} and EDSR~\cite{Lim_2017_CVPR_Workshops}.
    Soon afterward, residual dense blocks~\cite{zhang2018residual,tong2017image,huang2017densely}, channel attention~\cite{zhang2018image}, non-local networks~\cite{liu2018non} are subsequently incorporated into image super-resolution algorithms.
    Compared with image super-resolution, video super-resolution approaches are more focused on exploiting spatial and temporal information to restore high-quality images. 
    The common approach of video super-resolution is first performing frame temporal alignment using optical flow~\cite{xue2019video,liao2015video,kappeler2016video}, deformable convolution~\cite{tian2020tdan,wang2019edvr} or spatio-temporal network~\cite{caballero2017real,liu2017robust,sajjadi2018frame,kim2018spatio}, then conducting image synthesis that resembles single frame super-resolution.
    
    Recently, reference-based image super-resolution algorithms are widely used in the multi-camera imaging systems~\cite{cheng2020dual,zheng2018crossnet,paliwal2020deep,yang2020learning}, which transfers high-resolution textures from the given reference image to produce visually satisfactory results. 
    Paliwal~\textit{\etal}~\cite{paliwal2020deep} propose to use two video streams as inputs, including an auxiliary video with high-frame-rate, low resolution, and a main video with low-frame rate, high resolution.
    They use CNNs to compute the optical flows between missing frames and the two existing frames of the main stream by utilizing the content of the auxiliary video frames.
    Similar to Paliwal~\textit{\etal}~\cite{paliwal2020deep},~Cheng~\etal~\cite{cheng2020dual} propose a model for high spatio-temporal resolution video super-resolution.
    They also adopt an auxiliary stream and a main stream, but they just use one high-resolution reference frame of the main stream and the low-resolution frames of the auxiliary video to synthesize the high-resolution images.
    Besides, the optical flows are estimated only using consecutive auxiliary frames in Paliwal~\textit{\etal}~\cite{paliwal2020deep}'s model, while Cheng~\textit{\etal}~\cite{cheng2020dual} use the up-scaled auxiliary frames as well as the reference image to perform motion estimation, similar to the setting of CrossNet~\cite{zheng2018crossnet}.
    However, those method does not consider the situation when the low-resolution is too small to support optical flow estimation, that will dramatically degrade the quality of synthesized frames.

\Paragraph{Video Frame Prediction.} 
    Recent video frame prediction or extrapolation algorithms are using pixel-based~\cite{denton2018stochastic,byeon2018contextvp,lotter2016deep}, motion-based~\cite{liu2017video,liu2018future}, and joint motion-pixel based schemes~\cite{gao2019disentangling,hao2018controllable,reda2018sdc}.
    Pixel-based methods that generate pixels from scratch typically use implicit motion representations, which often leads to blurry effects.
    While the motion-based approaches produce sharp content, but the occlusion areas are often erroneous.
    The joint motion-pixel based algorithms such as DPG~\cite{gao2019disentangling} gates the output of pixel predictions from a flow predictor for non-occluded regions and from a context encoder for occluded regions, respectively.
    However, most motion-based video prediction methods predict the future optical flow by inputting multiple historical frames using a CNN (such as a U-Net) directly.
    Without explicit flow modeling, the error-prone predicted flow will degrade the quality of synthesized images.

\begin{figure*}[t]
	\footnotesize
	%\tiny
	%\scriptsize
	\centering
	\renewcommand{\tabcolsep}{2.0pt} % adjust horizontal space
	\renewcommand{\arraystretch}{1.2} % adjust vertical space
	\begin{tabular}{cccccc}
                % \multicolumn{2}{c}{\includegraphics[width=0.5\linewidth, height=0.2\linewidth]{preface-figure/blurry_overlap/blurry.png}} &
                \includegraphics[width=0.16\linewidth]{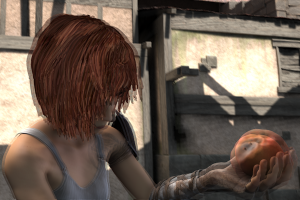} &
                \includegraphics[width=0.16\linewidth]{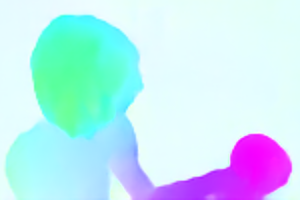} &
                 \includegraphics[width=0.16\linewidth]{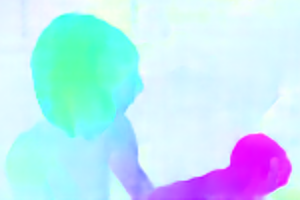} &
				  \includegraphics[width=0.16\linewidth]{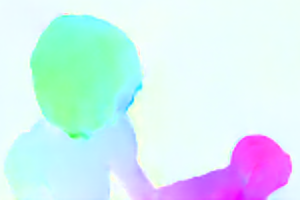} &
			 \includegraphics[width=0.16\linewidth]{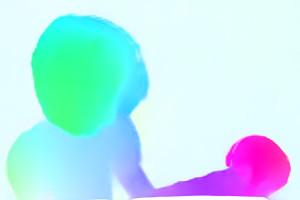} &
			 \includegraphics[width=0.16\linewidth]{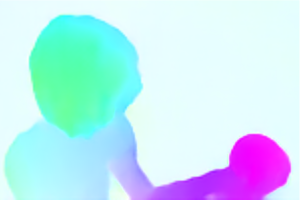} 
			    \\
			    (a) overlayed inputs &
				(b) estimated ($s=4$) &
				(c) estimated ($s=8$) &
				(d) propagated &
				(e) warped &
				(f) fused 
				\\
	\end{tabular}
	\vspace{-5pt}
	\caption{
        \textbf{Visualizations of optical flow outputs of prediction, propagation and estimation schemes on the Sintel dataset~\cite{sintel}.}
	}
	\label{fig:flow_acc_scale} %% label for entire figure
	\vspace{-5pt}
\end{figure*}

\section{Proposed Model}
    
    In this section, we introduce the proposed \textbf{P}rediction-\textbf{ASS}istant Network (PASS-Net).
    The architecture of the proposed model is shown in~\figref{fig:framework}.

    \subsection{Flow Prediction}
    We first consider the case where the future frame is lost (i.e., not received before the deadline).
    Given $n+1$ previously received HR frames $\mathbf{I}_{-n},\dots, \mathbf{I}_{-1}, \mathbf{I}_{0}$,~the goal of flow propagation is obtaining the optical flow from the current frame $\mathbf{I_0}$ to the future frame $\mathbf{\mathbf{I}}_t$.
    In this work, we set $n$ equals to two.
    To predict the future optical flow, a straightforward approach is \textit{flow warping}.
    The flow warping operation works effectively in optical flow estimation algorithms~\cite{ren2019fusion}.
    We first compute the optical flow fields between frame $\mathbf{I}_{-1}$ and $\mathbf{I}_0$, denoted by $\mathbf{F}_{-1\rightarrow 0}$ and $\mathbf{F}_{0 \rightarrow -1}$.
    Then we backward warp $\mathbf{F}_{-1 \rightarrow 0}$ with $\mathbf{F}_{0 \rightarrow -1}$ to approximate the flow from time $0$ to $1$.
    \begin{equation}
        \mathbf{\hat{F}}_{0 \rightarrow 1} = \mathcal{W}(\mathbf{F}_{-1 \rightarrow 0};~\mathbf{F}_{0 \rightarrow -1}),
    \end{equation}
    where $\mathcal{W}(\mathbf{x}; \mathbf{f})$ refers to the result of back-warping the input $\mathbf{x}$ based on the optical flow field $\mathbf{f}$.
    Existing models~\cite{jiang2018super,bao2019depth, niklaus2018context} usually assume a uniform motion between consecutive frames, the optical flow from $0$ to $t$ can be derived as:
    \begin{equation}
        \label{eq:flow1}
        \mathbf{\hat{F}}_{0 \rightarrow t} = t \cdot \mathbf{\hat{F}}_{0 \rightarrow 1}.
    \end{equation}

    However, the linear assumption may not hold true for the complex non-uniform motion, especially when the future frame $\mathbf{I}_t$ is lost.
    In contrast, we take inspirations from the work by Xu~\textit{\etal}~\cite{xu2019quadratic} where higher-order information is considered for more accurate motion modeling.
    We extend their method from flow interpolation into \textit{flow propagation}.
    
    Given three previously received frames $\mathbf{I}_{-2}$, $\mathbf{I}_{-1}$ and $\mathbf{I}_{0}$, we first compute two optical flow fields $\mathbf{F}_{0 \rightarrow -1}$ and $\mathbf{F}_{0 \rightarrow -2}$. 
    Under the assumption of the uniform acceleration~\cite{xu2019quadratic}, these optical flow fields can be modeled using $\mathbf{F}_{0 \rightarrow m} =\mathbf{v}_0 \cdot m + \frac{1}{2}\cdot \mathbf{a} \cdot m^2$, where $m \in [-1, -2]$, $\mathbf{v}_0$ and $\mathbf{a}$ denote speed and acceleration respectively.
    Eliminating $\mathbf{v}_0$ and $\mathbf{a}$ (more details can be found in the supplementary materials), the propagated optical flow field $\mathbf{\tilde{F}}_{0 \rightarrow t}$ can be derived as:
    \begin{equation}
        \label{eq:flow2}
        \mathbf{\tilde{F}}_{0 \rightarrow t} = 0.5t(t+1)\mathbf{F}_{0 \rightarrow -2} - t(t+2)\mathbf{F}_{0 \rightarrow -1}.
    \end{equation}
    With \eqnref{eq:flow1} and (\ref{eq:flow2}), we have two candidate representations, namely  $\mathbf{\tilde{F}}_{0 \rightarrow t}$ and $\mathbf{\hat{F}}_{0 \rightarrow t}$ for the future optical flow.

    We compare the warped and propagated flows with the ground truth at every pixel using the Sintel dataset~\cite{sintel} with $t=1$.
    We test the optical flow fields for two flow methods, FlowNet2-S~\cite{IMKDB17} and PWC-Net~\cite{sun2018pwc}.
    As shown in~\tabref{tab:flow_prop}, in most cases, the warped flows are more accurate than the propagated flows using the uniform acceleration assumption.
    However, the propagated can also benefit the flow prediction in complex motion cases.
    For the FlowNet2-S based backbone model, the error of initial flow estimation will accumulate in the propagation process.
    Thus, selecting an accurate backbone of flow estimation is significant for flow prediction.
    In this paper, we select the state-of-the-art PWC-Net~\cite{sun2018pwc} as our optical flow estimator.

    \subsection{Flow Estimation}
    
    We then consider the case where the received current frame is in low-quality.
    We use the resolution degradation as the quality degradation model.
    \begin{equation}
    \mathbf{I}_t^{\downarrow s} = \mathcal{D} (\mathbf{I}_t ; s),
    \end{equation}
    where $\mathcal{D}$ denotes a degradation mapping function, e.g. bicubic sampling, $\mathbf{I}_t$ is the corresponding high-resolution counterpart and $s$ represents the scaling factor.
    Our goal is to estimate optical flow field $\mathbf{\bar{F}}_{0 \rightarrow t}$.
    \begin{equation}\label{eq1}
        \mathbf{\bar{F}}_{0 \rightarrow t} = \mathbf{E}(\mathbf{I}_{-n}, \dots, \mathbf{I}_0;~\mathbf{I}_t^{\downarrow s}).
    \end{equation}
    In this paper, we only use one historical frame $\mathbf{I}_0$, the~\eqnref{eq1} can be reformulated to $\mathbf{E}(\mathbf{I}_0;~\texttt{bic}(\mathbf{I}_t^{\downarrow s}))$, $\mathbf{E}$ is a flow estimation function, e.g. PWC-Net~\cite{sun2018pwc}, $\texttt{bic}(\cdot)$ refers to bicubic up-scaling function.

\begin{figure}[t]
	\footnotesize
    % \scriptsize

	\centering
% 	\renewcommand{\tabcolsep}{1.0pt} % adjust horizontal space
% 	\renewcommand{\arraystretch}{1.0} % adjust vertical space
% 	\begin{tabular}{c}
	\includegraphics[width= 0.47\textwidth]{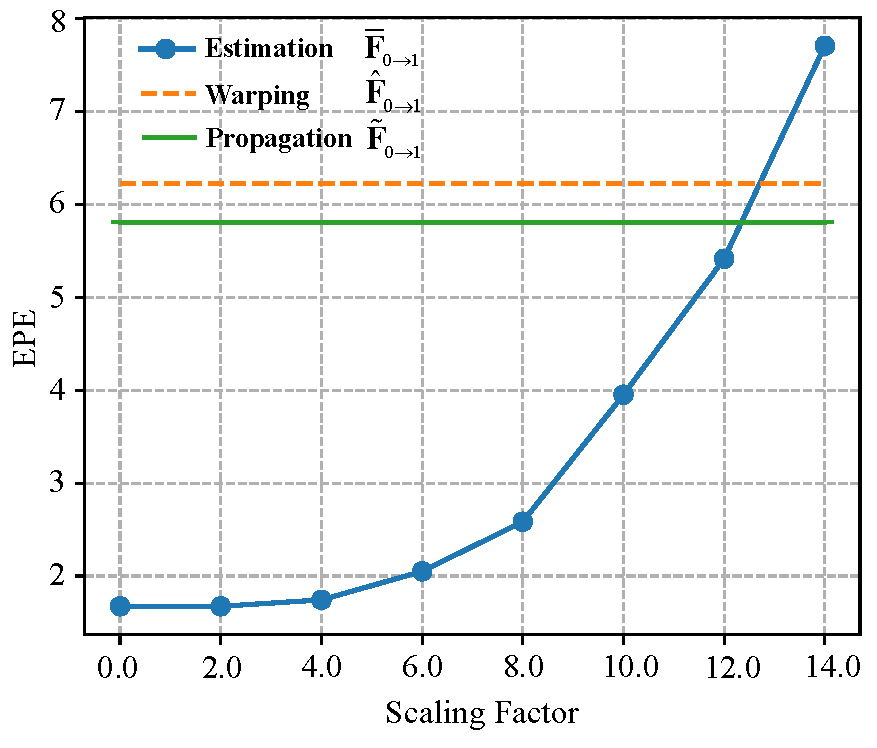}
	
% \end{tabular}
 	\vspace{-5pt}
	\caption{
		\textbf{Analysis of the accuracy of optical flow estimation considering scaling factor on the Sintel~\cite{sintel} dataset.}
		We estimate the optical flow $\mathbf{E}(\mathbf{I}_0;~\texttt{bic}(\mathbf{I}_t^{\downarrow s}))$, where $\mathbf{E}$ refers to PWC-Net~\cite{sun2018pwc}, $\texttt{bic}(\mathbf{I}_1^{\downarrow s})$ is the bicubic interpolated frame.
	}
 	\vspace{-10pt}
	\label{fig:acc_scale} 
\end{figure}

    We conduct ablation studies to investigate when the accuracy of propagated flow $\mathbf{\tilde{F}}_{0 \rightarrow t}$ or $\mathbf{\hat{F}}_{0 \rightarrow t}$ is higher than directly estimation $\mathbf{\bar{F}}_{0 \rightarrow t}$ by adjusting the scaling factor $s \in \{0,2,4,6,8,10,12,14\}$ with $t = 1$.
    As shown in~\figref{fig:acc_scale}, with the increase of scaling factor $s$, the resolution of frame $\mathbf{I}_t^{\downarrow s}$ decreases, which means the amount of pixel information decreases, indicating a quality degradation.
    When the scaling factor $s$ exceeds 12, the flow prediction performs better than flow estimation cause the received frame $\mathbf{I}_t^{\downarrow s}$ lost too much information.
    
    \subsection{Flow Fusion}

    Our model works in two cases, the lost case and the lossy case.
    For the lost case, we obtain the propagated flow $\mathbf{\tilde{F}}_{0 \rightarrow t}$ and warped flow $\mathbf{\hat{F}}_{0 \rightarrow t}$, as shown in~\figref{fig:framework}.
    As shown in~\figref{fig:flow_acc_scale}, the warped flow tends to miss parts of the pixels around the image boundary because of back-warping and the propagated flow still has some ringing artifacts around object edges.
    For the lossy case, besides the propagated flow and warped flow, we also obtain the estimated flow $\mathbf{\bar{F}}_{0 \rightarrow t}$.
    However, with the increase of scaling factor $s$, the accuracy of the estimated flow decreases rapidly, and the flow field presents large artifacts across the whole figure, as shown in~\figref{fig:flow_acc_scale}.
    
    In this section, we propose a flow fusion module to refine those candidate flows. 
    We refine the flow field using a U-Net.
    We concatenate the propagated flow $\mathbf{\tilde{F}}_{0 \rightarrow t}$, warped flow $\mathbf{\hat{F}}_{0 \rightarrow t}$ and estimated flow $\mathbf{\bar{F}}_{0 \rightarrow t}$ if available as well as image $\mathbf{I}_0$ as inputs to the fusion module.
    We formulate the initial flow fusion process of the lost case as follows:
    \begin{equation}
        \mathbb{\hat{F}}_{0 \rightarrow t} = \mathcal{F}(\mathbf{\tilde{F}}_{0 \rightarrow t}, \mathbf{\hat{F}}_{0 \rightarrow t}, \mathbf{I}_0),
    \end{equation}
    and the lossy case as follows:
    \begin{equation}
        \mathbb{\hat{F}}_{0 \rightarrow t} = \mathcal{F}(\mathbf{\tilde{F}}_{0 \rightarrow t}, \mathbf{\hat{F}}_{0 \rightarrow t}, \mathbf{\bar{F}}_{0 \rightarrow t}, \mathbf{I}_0),
    \end{equation}
    where image $\mathbf{I}_0$ is used to provide high-frequency details, and $\mathcal{F}$ refers to the flow fusion module.
    Moreover, we learn the residual optical flow using the U-net, which can be added to the initial estimated optical flow for refinement.
    We formulate the refinement process as follows:
    \begin{equation}
        \mathbb{F}_{0 \rightarrow t}(\mathbf{u}) = \mathbb{\hat{F}}_{0 \rightarrow t} (\mathbf{u} + \sigma(\mathbf{u})) + \mathbf{r}(\mathbf{u}),
    \end{equation}
    where $\sigma(\mathbf{u})$ is the learned sampling offset of pixel $\mathbf{u}$, the $\sigma(\mathbf{u})$ is constrained to $[-k,k]$ by $k \times \texttt{tanh}(\cdot)$ as the activation function to achieve a local receptive field of $2k+1$, and the $\mathbf{r}(\mathbf{u})$ is the learned residual flow.

    \subsection{Frame Synthesis}
    
    We align the input reference frame $\mathbf{I}_0$ to the target frame at time $t$ according to the back-projected fused flow $\mathbb{F}_{t \rightarrow 0}$ to synthesize an initial frame.
    We use the mean back-propagation process which works effectively in frame interpolation task~\cite{xue2019video, bao2018memc}:
    \begin{equation}
        \mathbb{F}_{t \rightarrow 0}(\mathbf{u}) = \frac{
            -\sum_{\mathbf{x}+ \mathbb{F}_{0 \rightarrow t}(\mathbf{u}) \in \mathcal{N}(\mathbf{u})} w(||\mathbf{x} - \mathbf{u}||_2)\mathbb{F}_{0 \rightarrow t}(\mathbf{x})
        }{
        \sum_{\mathbf{x}+ \mathbb{F}_{0 \rightarrow t}(\mathbf{u}) \in \mathcal{N}(\mathbf{u})}w(||\mathbf{x} - \mathbf{u}||_2)
        },
    \end{equation}
    where $\mathbf{u}$ is a pixel coordinate on $\mathbf{I}_t$, $\mathcal{N}(\mathbf{u})$ is the neighborhood of $\mathbf{u}$, and $w(d) = e^{-d^2/\sigma^2}$ is the Gaussian weight for flow.
    At last, we synthesize the result $\mathbf{I}_t$ by concatenate flow, up-scaled frame $\mathbf{I}_{t}^{\downarrow s}$ if available, using the image synthesis network $\phi$.
    The synthesis process based on back-warping operation $\mathcal{W}$ is given by:
     \begin{equation}
        \mathbf{I}_t = \phi \Big( 
            \mathcal{W}(\mathbf{I}_0;
              \mathbb{F}_{t \rightarrow 0}), \texttt{bic}(\mathbf{I}_t^{\downarrow s}), \mathbb{F}_{t \rightarrow 0} ; \Theta\Big).
    \end{equation}
    We use the U-Net as the image synthesis network $\phi$ with the trainable parameter set $\Theta$.
    
    \subsection{Implementation Details}
    
\begin{figure}[t]
	\footnotesize
    % \scriptsize

	\centering
% 	\renewcommand{\tabcolsep}{1.0pt} % adjust horizontal space
% 	\renewcommand{\arraystretch}{1.0} % adjust vertical space
% 	\begin{tabular}{c}
	\includegraphics[width= 0.475\textwidth]{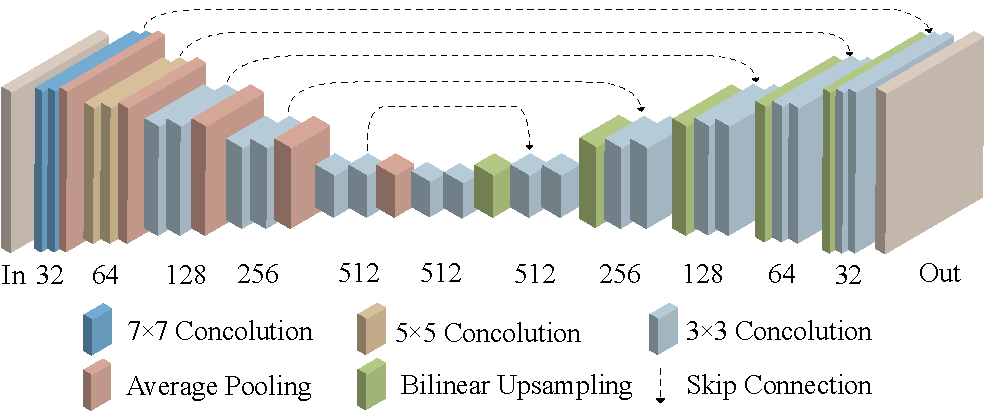}
	
% \end{tabular}
 	\vspace{-5pt}
	\caption{
		\textbf{Architecture of the U-Net used in Flow Fusion module and Frame Synthesis Module.} 
	}
 	\vspace{-5pt}
	\label{fig:u-net} 
\end{figure}

    \Paragraph{Network Components.} 
    As shown in~\figref{fig:framework}, the model consists of four neural network components.
    We adopt the PWC-Net~\cite{sun2018pwc} as our flow estimation network.
    The flow prediction module also re-uses this estimation network multiple times.
    Except for the PWC-Net, the flow prediction module has no learnable parameters.
    The flow fusion module and frame synthesize module are both U-Net, the only difference is inputs and outputs, as shown in~\figref{fig:u-net}.

    \Paragraph{Loss Function.} 
    We denote the synthesized frame as $\mathbf{I}_t$, and the ground-truth image is denoted by $\mathbf{I}_t^{\text{GT}}$.
    We use the following pixel loss function to train our model:
     \begin{equation}\label{pixelloss}
            \mathcal{L} =  \sum_{\mathbf{u}}  \rho \left( \mathbf{I}_t (\mathbf{u}) - \mathbf{I}_{t}^{\text{GT}} (\mathbf{u}) \right),
    \end{equation}
    where $\rho({x}) = \sqrt{{x}^2+\epsilon^2}$ is the Charbonnier penalty function~\cite{charbonnier1994two}.
    We set the constant $\epsilon$ to $1e^{-6}$.

    \Paragraph{Training Dataset.} 
    We train the proposed model \textit{only} on the Vimeo90K-Septuplet~\cite{xue2019video} training set.
    The septuplet dataset consists of 91,701 7-frame sequences with a fixed resolution of ${448 \times 256}$, which are extracted from 39000 selected video clips from Vimeo-90K.
    The low-resolution image distortion for the Vimeo90K dataset is generated by down-sampling the original high-resolution frames.
    We use the MATLAB function \texttt{imresize} with the \textit{bicubic} mode to accomplish the dataset generation.
    When training, we use the first three high-resolution frames as previously received frames and use the down-scaled fourth frame as the current low-resolution frame for the lossy case.
    For the lost case, we do not use any current frames.
    We augment the training data by horizontal and vertical flipping as well as randomly crop frames to a size of $256 \times 256$.
    
     \Paragraph{Training Strategy.}
     We use the AdaMax~\cite{kingma2014adam} optimizer with parameters $\beta_1=0.9$ and $\beta_2=0.999$ to train the model.
     The batch size is 8, and the initial learning rate is $1e^{-4}$. 
     We load the pre-trained weights of PWC-Net~\cite{sun2018pwc} when initializing and set the learning rate of PWC-Net to zero.
     We train the model for 80 epochs, then reduce the learning rate by a factor of 0.2 and fine-tune the PWC-Net for another 10 epochs.
     We train the model on an RTX-2080 Ti GPU card, which takes about five days to converge.
    
\section{Experimental Results}
    In this section, we first introduce the evaluation datasets and then conduct ablation studies to analyze the contribution of each proposed component.
    Then, we compare the proposed model with state-of-the-art algorithms for next frame synthesis.
    Finally, we analyze our model for the task of the next group of picture synthesis.
    
    \subsection{Evaluation Datasets and Metrics}
    
    We evaluate the proposed model on multiple public datasets with different image resolutions.
    The low-resolution frames are all generated using MATLAB function \texttt{imresize} with the \textit{bicubic} mode.
     
    \Paragraph{UCF101.}
    The UCF101 dataset~\cite{soomro2012ucf101} contains videos with variety of human actions.
    We use the evaluation set the same with DVF~\cite{liu2017video} to evaluate video prediction.

    \Paragraph{Vimeo90K.}
    There are $7824$ 7-frames in the test set of the Vimeo90K Septuplet dataset~\cite{xue2019video}.
    The image resolution of the test set is $448 \times 256$ pixels. 
    
    \Paragraph{Middlebury.}
    We use the Middlebury OTHER images as the evaluation dataset~\cite{baker2011database}.
    The image resolution of this dataset is around $640 \times 480$ pixels.
    
    \Paragraph{Vid4.}
    The Vid4~\cite{liu2013bayesian} dataset contains four video sequences: \textit{city}, \textit{walk}, \textit{calendar}, and \textit{foliage}.
    Each sequence in the Vid4 has at least 30 video frames with $720 \times 480$~pixels.
    
    \begin{table}[t]
  \caption{
     \textbf{Analysis of flow propagation and estimation using the Sintel~\cite{sintel} dataset.}
     For frame prediction, we take three previous frames $\mathbf{I}_{-2}, \mathbf{I}_{-1}, \mathbf{I}_{0}$ as inputs to \underline{propagate} or warp the optical flow of the next frame, denoted by \underline{$\mathbf{\tilde{F}}_{0 \rightarrow 1}$}, $\mathbf{\hat{F}}_{0 \rightarrow 1}$.
     For frame estimation, we use the current frame $\mathbf{I}_0$ and the next frame $\mathbf{I}_1$ of full-resolution to estimate optical flow $\mathbf{\bar{F}}_{0 \rightarrow 1}$.
     We compare the output flow with the ground truth at every pixel using end-point-error (EPE).
     The numbers in \first{red} and \second{blue} represent the best and second-best results.
  }
%   \vspace{-5pt}
  \label{tab:flow_prop}
  \footnotesize
  
  \renewcommand{\tabcolsep}{1.5pt} % adjust horizontal space
  \renewcommand{\arraystretch}{1.2} % adjust vertical space
\newcommand{\quantTit}[1]{\multicolumn{3}{c}{\scriptsize #1}}
    \newcommand{\quantSec}[1]{\scriptsize #1}
    \newcommand{\quantInd}[1]{\scriptsize #1}
    \newcommand{\quantVal}[1]{\scalebox{0.83}[1.0]{$ #1 $}}
    \newcommand{\quantBes}[1]{\scalebox{0.83}[1.0]{$\uline{ #1 }$}}
  \centering
  \begin{tabular}{lccc cccc c}
    \toprule
    
    % \multirow{2}{*}[-0.28em]{Method}   
    % %  & Runtime 
    % % & Parameters 
    %  &  \multicolumn{2}{c}{111}
    %   & \multicolumn{2}{c}{YouTube240} \\
    
    % %  \cmidrule(l{2pt}r{1pt}){2-3}
    % % \cmidrule(l{2pt}r{1pt}){4-5} 
    % \cmidrule(lr){2-3}
    % \cmidrule(lr){4-5}
    
    % % %\cmidrule(r){2-3} \cmidrule(r){4-5} \cmidrule(r){a-b}
    % & PSNR & SSIM  &PSNR & SSIM \\
    
    clips &  \quantSec{\text{market2}} & 
    \quantSec{\text{alley2}} & 
    \quantSec{\text{sleeping1}} & 
    \quantSec{\text{ambush6}} & 
    \quantSec{\text{alley1}} & 
    \quantSec{\text{temple2}} & 
    \quantSec{\text{cave4}}
    \\

    \midrule
    
    \multicolumn{8}{c}{ \quantSec{ Using FlowNet2-S~\cite{IMKDB17}}} 
    
    \\
    
    \midrule
        
        Propagation $\mathbf{\tilde{F}}_{0 \rightarrow 1}$ &
            
            \quantVal{2.586} & %market\_2
            \quantVal{1.653} & %alley\_2  
            \quantVal{0.919} & %sleeping\_1
            \quantVal{27.65} & %ambush\_6 
            \quantVal{1.684} & %alley\_1 
            \quantVal{6.937} & %temple\_2 
            \quantVal{9.938} % cave\_4
            
        \\
        Warping~~~~~~ $\mathbf{\hat{F}}_{0 \rightarrow 1}$ &
          
            \quantVal{\second{1.798}} & %market\_2
            \quantVal{\second{1.001}} & %alley\_2  
            \quantVal{\second{0.536}} & %sleeping\_1
            \quantVal{\second{25.46}} & %ambush\_6 
            \quantVal{\second{1.028}} & %alley\_1 
            \quantVal{\second{5.353}} & %temple\_2 
            \quantVal{\second{8.052}} % cave\_4
        \\
        
        Estimation~~ $\mathbf{\bar{F}}_{0 \rightarrow 1}$ &
          
            \quantVal{\first{1.306}} & %market\_2
            \quantVal{\first{0.770}} & %alley\_2  
            \quantVal{\first{0.528}} & %sleeping\_1
            \quantVal{\first{9.198}} & %ambush\_6 
            \quantVal{\first{0.818}} &%alley\_1 
            \quantVal{\first{3.059}} & %temple\_2 
            \quantVal{\first{5.227}} % cave\_4
            
        \\

        \midrule
        
        \multicolumn{8}{c}{ \quantSec{Using PWC-Net~\cite{sun2018pwc} }} 
        
        \\
        
        \midrule 
        
        Propagation $\mathbf{\tilde{F}}_{0 \rightarrow 1}$ &
          
            \quantVal{1.442} & %market\_2
            \quantVal{\second{0.561}} & %alley\_2  
            \quantVal{\second{0.287}} & %sleeping\_1
            \quantVal{\second{24.33}} & %ambush\_6 
            \quantVal{\second{0.587}} & %alley\_1 
            \quantVal{6.990} & %temple\_2 
            \quantVal{9.393} % cave\_4

        \\
        
        Warping~~~~~~ $\mathbf{\hat{F}}_{0 \rightarrow 1}$ &
             
            \quantVal{\second{1.297}}  & %market\_2
            \quantVal{0.757} & %alley\_2  
            \quantVal{0.139} & %sleeping\_1
            \quantVal{26.05}  & %ambush\_6 
            \quantVal{0.589} & %alley\_1 
            \quantVal{\second{4.171}} & %temple\_2 
            \quantVal{\second{7.643}} % cave\_4
        \\
        
        Estimation~~ $\mathbf{\bar{F}}_{0 \rightarrow 1}$ &

            \quantVal{\first{0.606}} & %market\_2
            \quantVal{\first{0.279}} & %alley\_2  
            \quantVal{\first{0.126}} & %sleeping\_1
            \quantVal{\first{5.320}} & %ambush\_6 
            \quantVal{\first{0.271}} & %alley\_1 
            \quantVal{\first{1.863}} &%temple\_2 
            \quantVal{\first{3.260}} % cave\_4
            
        \\
        
        % \midrule
        
        % Predict Fusion $\mathbb{{\hat{F}}}_{0 \rightarrow 1}$ &
        %      PWC-Net~\cite{sun2018pwc} &
            
        %     %market\_2
        %     %alley\_2  %sleeping\_1
        %     %ambush\_6 
        %     %alley\_1 
        %     %temple\_2 
        %     % cave\_4 
        %     \\
            
        % Estimate Fusion $\mathbb{{\bar{F}}}_{0 \rightarrow 1}$ &
        %      PWC-Net~\cite{sun2018pwc} &
            
        %     %market\_2
        %     %alley\_2  %sleeping\_1
        %     %ambush\_6 
        %     %alley\_1 
        %     %temple\_2 
        %     % cave\_4 
        %     \\

        \bottomrule

  \end{tabular}
    \vspace{-5pt}  
\end{table}

    \Paragraph{REDS.}
    REDS is a newly proposed high-quality video dataset~\cite{Nah_2019_CVPR_Workshops_SR}, which consists of 30 validation clips, and each with 100 consecutive frames at 720p.
    We extract every 7 frames to make the evaluation dataset.
    
    \Paragraph{HD.}
    The HD~\cite{bao2018memc} dataset consists of four $1920 \times 1080$, three $1280 \times 720$ and four $1280 \times 544$ videos.
    We use the $1920 \times 1080$ video for evaluation.
    We first convert the YUV video to regular RGB video using \texttt{FFmpeg}, then make multiple consecutive 7-frame groups to create the test set.

    \subsection{Model Analysis}
    
    We analyze the contribution of the two key components in the proposed network: the optical flow estimation module and the flow fusion module.

    \Paragraph{Optical Flow Estimation.}
    To analyze how well our approach performs with different correspondence estimates, we consider two state-of-the-art optical flow algorithms~\cite{sun2018pwc,IMKDB17}.
    As shown in~\tabref{tab:abla_component} (third  section), those methods all perform similarly well.
    And the PWC-Net based method performs better than FlowNet2-S based model because the PWC-Net learns a slightly accurate optical flow.
    Specifically, we fine-tune PWC-Net and achieve additional performance gain with this PWC-Net-ft.
    Thus, we use the fine-tuned PWC-Net for our algorithm.
    
     \begin{table}[t]
  \caption{
     \textbf{Ablation experiments to quantitatively analyze the effect of the different components of our model for next frame super-resolution.}
  }
%   \vspace{-1mm}
  \label{tab:abla_component}
  \footnotesize
% \scriptsize

  \renewcommand{\tabcolsep}{1.6pt} % adjust horizontal space
  \renewcommand{\arraystretch}{1.2} % adjust vertical space
  \newcommand{\quantTit}[1]{\multicolumn{3}{c}{\scriptsize #1}}
    \newcommand{\quantSec}[1]{\scriptsize #1}
    \newcommand{\quantInd}[1]{\scriptsize #1}
    \newcommand{\quantVal}[1]{\scalebox{0.83}[1.0]{$ #1 $}}
    \newcommand{\quantBes}[1]{\scalebox{0.83}[1.0]{$\uline{ #1 }$}}
    
  \centering
  \begin{tabular}{lcccccccc}
    \toprule

    \multirow{4}{*}[-0.28em]{Method}   
     & \multicolumn{4}{c}{Vimeo90K~\cite{xue2019video}}
     & \multicolumn{4}{c}{Middlebury~\cite{baker2011database}}

     \\
        \cmidrule(lr){2-5}
        \cmidrule(lr){6-9}

     & \multicolumn{2}{c}{ 4 $\times$}
     & \multicolumn{2}{c}{ 8 $\times$} 
 
     & \multicolumn{2}{c}{ 4 $\times$}
     & \multicolumn{2}{c}{ 8 $\times$}

     \\
    
        \cmidrule(lr){2-3}
        \cmidrule(lr){4-5}
        \cmidrule(lr){6-7}
        \cmidrule(lr){8-9}

    &  \quantSec{PSNR} & \quantSec{SSIM}  & \quantSec{PSNR} & \quantSec{SSIM}  
     & \quantSec{PSNR} & \quantSec{SSIM}  & \quantSec{PSNR} & \quantSec{SSIM} 

    \\
    
    \midrule
        Bicubic 
              
              & \quantVal{29.77} &  \quantVal{0.903}
              & \quantVal{25.79} &  \quantVal{0.824}
              
              & \quantVal{27.71}  & \quantVal{0.858}
              & \quantVal{23.82}  & \quantVal{0.763}
    \\

        \midrule
        
         PA - FlowNet2-S
              
              & \quantVal{39.79} & \quantVal{0.974}
              & \quantVal{37.84} & \quantVal{0.961}
              
              & \quantVal{35.04} & \quantVal{0.948}
              & \quantVal{\second{32.16}} & \quantVal{\second{0.936}}
              
              \\
        
        PA - PWC-Net
              
              & \quantVal{\second{41.01}} & \quantVal{\second{0.978}}
              & \quantVal{\second{37.98}} & \quantVal{\second{0.964}}
              
              & \quantVal{\second{35.18}} & \quantVal{\second{0.952}}
              & \quantVal{{32.15}} & \quantVal{\second{0.936}}
              
              \\
        
        PA - PWC-Net-ft
              
              & \quantVal{\first{41.72}} & \quantVal{\first{0.989}}
              & \quantVal{\first{38.81}} & \quantVal{\first{0.982}}
              
              & \quantVal{\first{36.40}} & \quantVal{\first{0.967}}
              & \quantVal{\first{33.66}} & \quantVal{\first{0.948}}
              
              \\
        
        \midrule
        
        PA - \textit{only f-estimation}
              
              & \quantVal{40.02} & \quantVal{0.968}
              & \quantVal{36.71} & \quantVal{0.956}
              
              & \quantVal{35.45} & \quantVal{0.947}
              & \quantVal{32.84} & \quantVal{0.926}
              
              \\

               PA - \textit{w/ f-propagation}
              
              & \quantVal{40.45} & \quantVal{0.970}
              & \quantVal{37.11} & \quantVal{0.962}

              & \quantVal{36.14} & \quantVal{0.959}
              & \quantVal{33.22} & \quantVal{0.932}

              \\
        
                PA -\textit{ w/ f-warping}
              & \quantVal{\second{40.46}} & \quantVal{\second{0.970}}
              & \quantVal{\second{37.41}} & \quantVal{\second{0.969}}

              & \quantVal{\second{36.22}} & \quantVal{\second{0.960}}
              & \quantVal{\second{33.43}} & \quantVal{\second{0.935}}
   
              \\
             
        PA
              
              & \quantVal{\first{41.72}} & \quantVal{\first{0.989}}
              & \quantVal{\first{38.81}} & \quantVal{\first{0.982}}
              
              & \quantVal{\first{36.40}} & \quantVal{\first{0.967}}
              & \quantVal{\first{33.66}} & \quantVal{\first{0.948}}
              
              \\
              
        \bottomrule

  \end{tabular}
    \vspace{-5pt}  
\end{table}

     \begin{figure*}[!th]
	\footnotesize
% 	\tiny 
	\scriptsize        
	\centering
	\renewcommand{\tabcolsep}{0.25pt} % adjust horizontal space
	\renewcommand{\arraystretch}{1.0} % adjust vertical space
	\begin{tabular}{ccccc ccccc}

% \multirow{3}{*}[10.10em]{\includegraphics[width=0.22\linewidth]{compare_figures/middlebury/gt_s.png}} &

\includegraphics[width=0.099\linewidth]{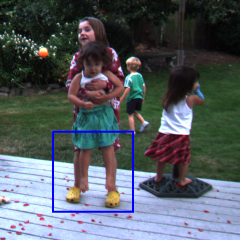}&
\includegraphics[width=0.099\linewidth]{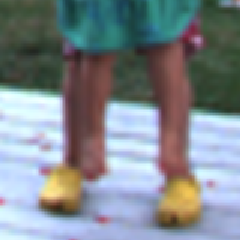}&
\includegraphics[width=0.099\linewidth]{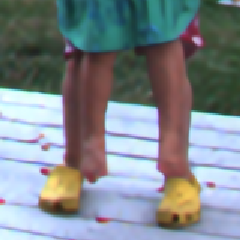}&
\includegraphics[width=0.099\linewidth]{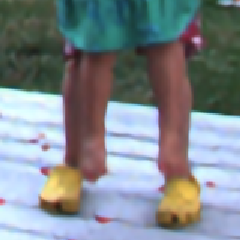}&
\includegraphics[width=0.099\linewidth]{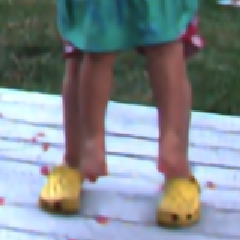} &

\includegraphics[width=0.099\linewidth]{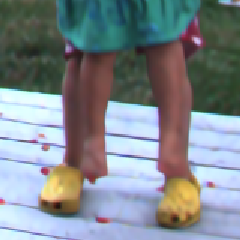}  &
\includegraphics[width=0.099\linewidth]{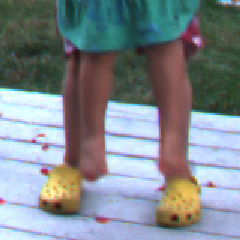}&
\includegraphics[width=0.099\linewidth]{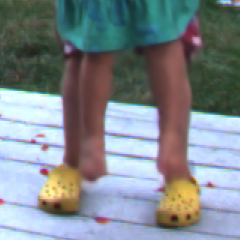} &
\includegraphics[width=0.099\linewidth]{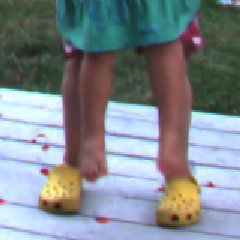} &
\includegraphics[width=0.099\linewidth]{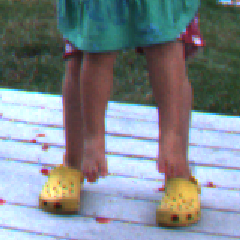}
\\

\includegraphics[width=0.099\linewidth]{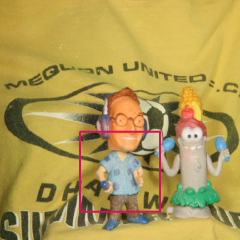}&
\includegraphics[width=0.099\linewidth]{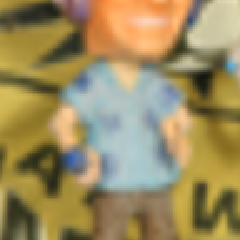}&
\includegraphics[width=0.099\linewidth]{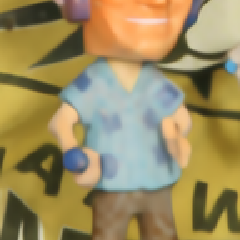}&
\includegraphics[width=0.099\linewidth]{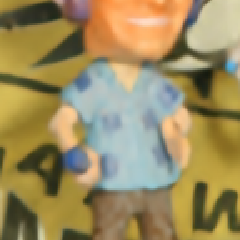}&
\includegraphics[width=0.099\linewidth]{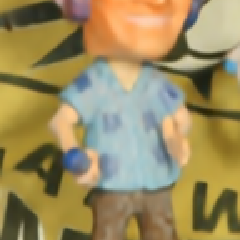} &

\includegraphics[width=0.099\linewidth]{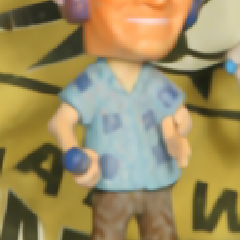}  &
\includegraphics[width=0.099\linewidth]{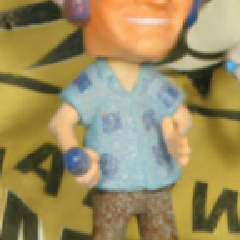}&
\includegraphics[width=0.099\linewidth]{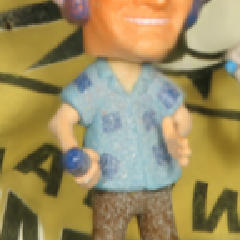} &
\includegraphics[width=0.099\linewidth]{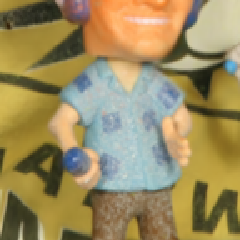} &
\includegraphics[width=0.099\linewidth]{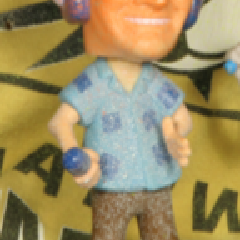}
\\

GT & 
Bicubic &
EDSR &
TOFlow &
DUF &
EDVR &
CrossNet &
AWnet &
Ours &
GT 

\\

\end{tabular}
	\vspace{-5pt}
	\caption{
       \textbf{Visual comparisons on the Middlebury set~\cite{baker2011database}.}
	}
	\label{fig:compare_middlebury} %% label for entire figure
	\vspace{-10pt}
\end{figure*}

     \begin{figure*}[!th]
	\footnotesize
% 	\tiny 
	\scriptsize        
	\centering
	\renewcommand{\tabcolsep}{0.20pt} % adjust horizontal space
	\renewcommand{\arraystretch}{1.0} % adjust vertical space
	\begin{tabular}{ccccc ccccc}

% \multirow{3}{*}[10.10em]{\includegraphics[width=0.22\linewidth]{compare_figures/middlebury/gt_s.png}} &

\includegraphics[width=0.099\linewidth]{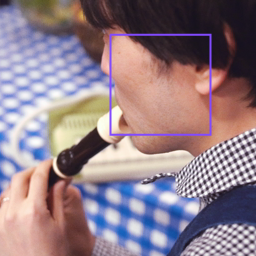}&
\includegraphics[width=0.099\linewidth]{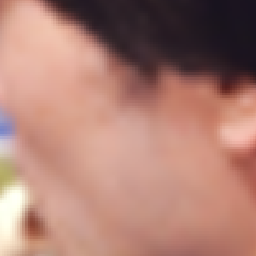}&
\includegraphics[width=0.099\linewidth]{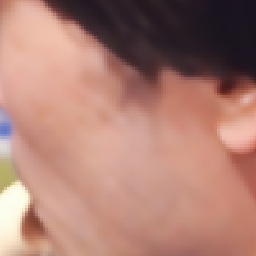}&
\includegraphics[width=0.099\linewidth]{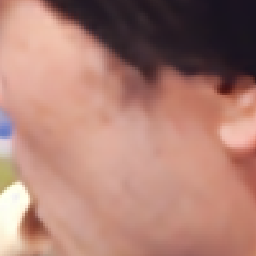}&
\includegraphics[width=0.099\linewidth]{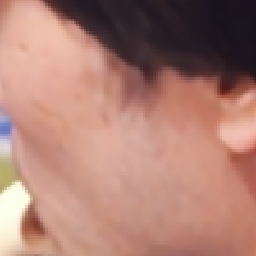} &

\includegraphics[width=0.099\linewidth]{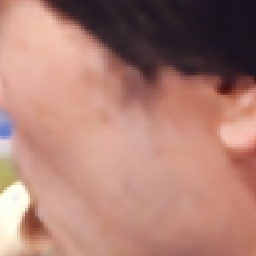}  &
\includegraphics[width=0.099\linewidth]{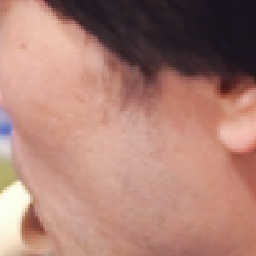}&
\includegraphics[width=0.099\linewidth]{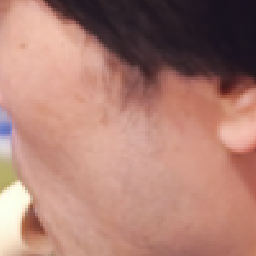} &
\includegraphics[width=0.099\linewidth]{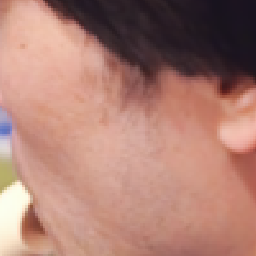} &
\includegraphics[width=0.099\linewidth]{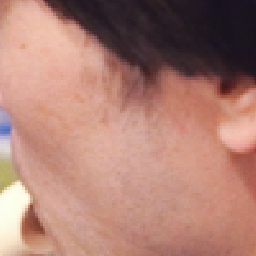}
\\

\includegraphics[width=0.099\linewidth]{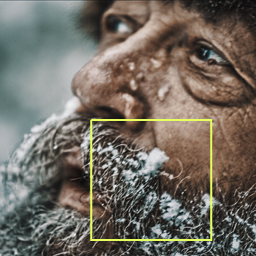}&
\includegraphics[width=0.099\linewidth]{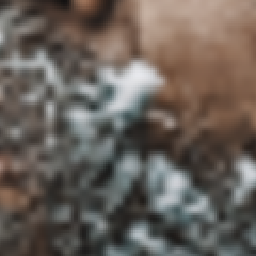}&
\includegraphics[width=0.099\linewidth]{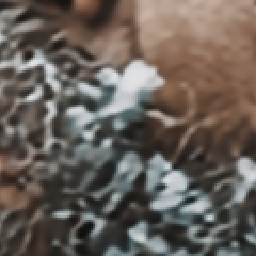}&
\includegraphics[width=0.099\linewidth]{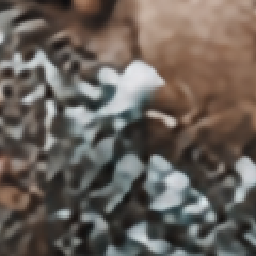}&
\includegraphics[width=0.099\linewidth]{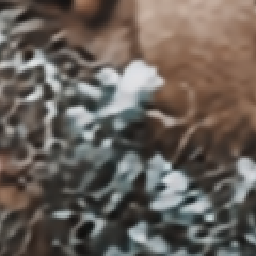} &

\includegraphics[width=0.099\linewidth]{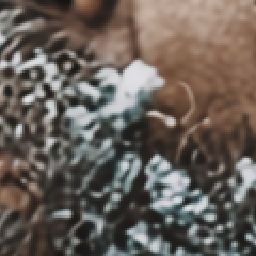}  &
\includegraphics[width=0.099\linewidth]{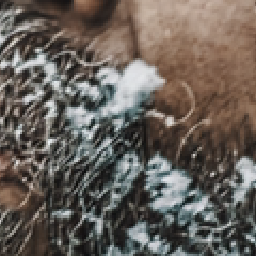}&
\includegraphics[width=0.099\linewidth]{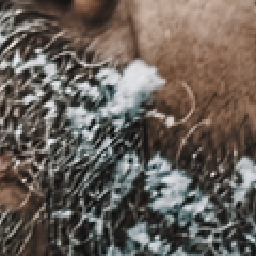} &
\includegraphics[width=0.099\linewidth]{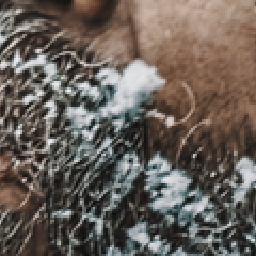} &
\includegraphics[width=0.099\linewidth]{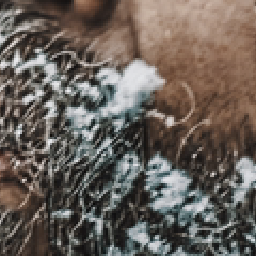}
\\

GT & 
Bicubic &
EDSR &
TOFlow &
DUF &
EDVR &
CrossNet &
AWnet &
Ours &
GT 

\\

\end{tabular}
	\vspace{-5pt}
	\caption{
       \textbf{Visual comparisons on the Vimeo90K set~\cite{xue2019video}.}
	}
	\label{fig:compare_vimeo} %% label for entire figure
	\vspace{-10pt}
\end{figure*}

     \begin{figure*}[!th]
	\footnotesize
% 	\tiny 
	\scriptsize        
	\centering
	\renewcommand{\tabcolsep}{0.12pt} % adjust horizontal space
	\renewcommand{\arraystretch}{1.0} % adjust vertical space
	\begin{tabular}{ccccc ccccc}

% \multirow{3}{*}[10.10em]{\includegraphics[width=0.22\linewidth]{compare_figures/middlebury/gt_s.png}} &

\includegraphics[width=0.099\linewidth]{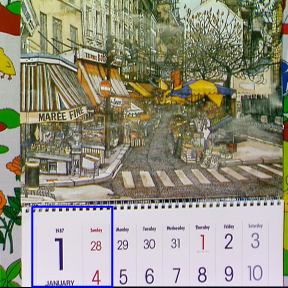}&
\includegraphics[width=0.099\linewidth]{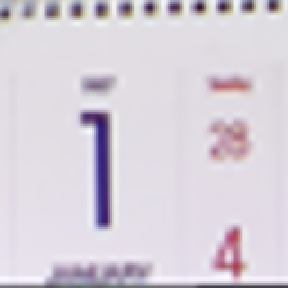}&
\includegraphics[width=0.099\linewidth]{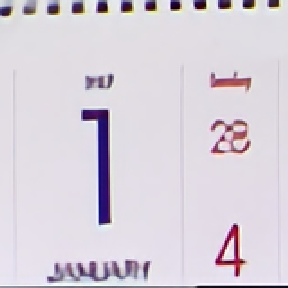}&
\includegraphics[width=0.099\linewidth]{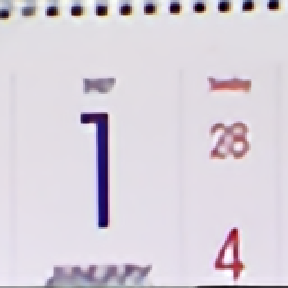}&
\includegraphics[width=0.099\linewidth]{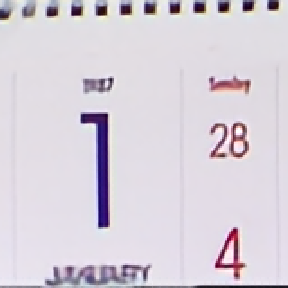} &

\includegraphics[width=0.099\linewidth]{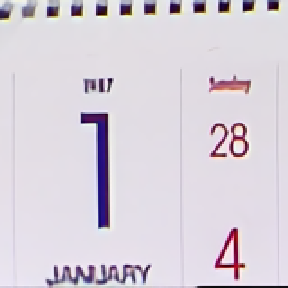}  &
\includegraphics[width=0.099\linewidth]{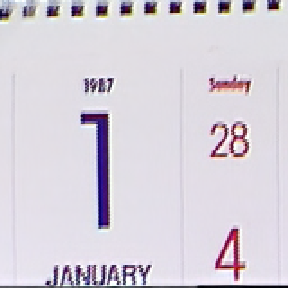}&
\includegraphics[width=0.099\linewidth]{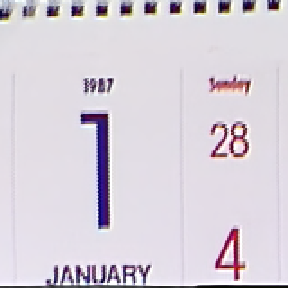} &
\includegraphics[width=0.099\linewidth]{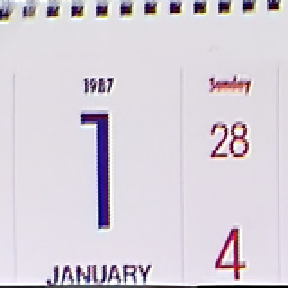} &
\includegraphics[width=0.099\linewidth]{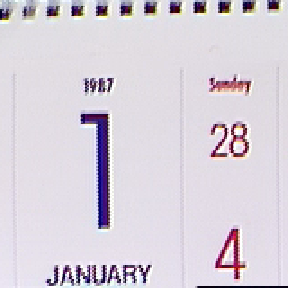}
\\

GT & 
Bicubic &
EDSR &
TOFlow &
DUF &
EDVR &
CrossNet &
AWnet &
Ours &
GT 

\\

\end{tabular}
	\vspace{-5pt}
	\caption{
       \textbf{Visual comparisons on the Vid4 set~\cite{liu2013bayesian}.}
	}
	\label{fig:compare_vid4} %% label for entire figure
	\vspace{-10pt}
\end{figure*}

     \begin{figure*}[!th]
	\footnotesize
% 	\tiny 
	\scriptsize        
	\centering
	\renewcommand{\tabcolsep}{0.12pt} % adjust horizontal space
	\renewcommand{\arraystretch}{1.0} % adjust vertical space
	\begin{tabular}{ccccc ccccc}

% \multirow{3}{*}[10.10em]{\includegraphics[width=0.22\linewidth]{compare_figures/middlebury/gt_s.png}} &

\includegraphics[width=0.099\linewidth]{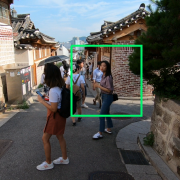}&
\includegraphics[width=0.099\linewidth]{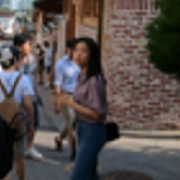}&
\includegraphics[width=0.099\linewidth]{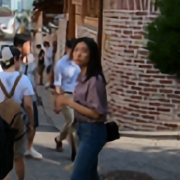}&
\includegraphics[width=0.099\linewidth]{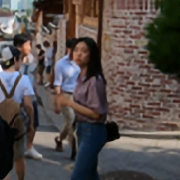}&
\includegraphics[width=0.099\linewidth]{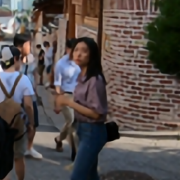} &

\includegraphics[width=0.099\linewidth]{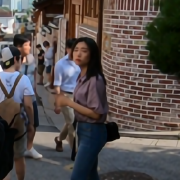}  &
\includegraphics[width=0.099\linewidth]{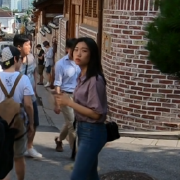}&
\includegraphics[width=0.099\linewidth]{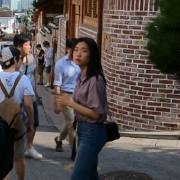} &
\includegraphics[width=0.099\linewidth]{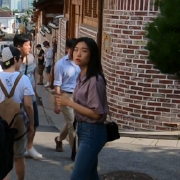} &
\includegraphics[width=0.099\linewidth]{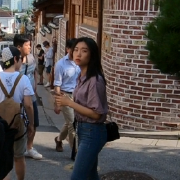}
\\

GT & 
Bicubic &
EDSR &
TOFlow &
DUF &
EDVR &
CrossNet &
AWnet &
Ours &
GT 

\\

\end{tabular}
	\vspace{-5pt}
	\caption{
       \textbf{Visual comparisons on the REDS set~\cite{Nah_2019_CVPR_Workshops_SR}.}
	}
	\label{fig:compare_reds} %% label for entire figure
	\vspace{-10pt}
\end{figure*}

    \Paragraph{Frame Prediction Assistant.}
    Our video super-resolution model uses the previous high-resolution frames and current low-resolution frames to synthesize high-quality frames at the current time.
    We adopt warped flow and propagated flow to help the flow fusion module to produce an accurate future flow, as shown in~\figref{fig:framework}.
    We train four versions of our model, including only using flow estimation~(PA~-~\textit{only-f-estimation}), without using flow propagation~(PA~-~\textit{w/f-propagation}), without using flow warping~(PA~-~\textit{w/f-warping}), and using all of them (PA).
    As shown in~\tabref{tab:abla_component} (fourth section), the model using the propagated flow and warped frame performs better than the algorithm only relies on flow estimation.
    Furthermore, when the current low-resolution frame is lost, we conduct experiments to verify the contribution of flow propagation for video frame prediction.
    Our model for frame prediction using the propagated frame works better than the one without using it, as shown in~\tabref{tab:compare_prediction}.
    Thus, the propagated flow and warped flow provide a foundation for the lost case, and the lossy case.

    \begin{table*}[t]
  \caption{
     \textbf{Quantitative comparisons of various state-of-the-art methods for next frame super-resolution.}
  }
%   \vspace{-5pt}
  \label{tab:compare_next_frame_sr}
  \footnotesize
% \scriptsize

  \renewcommand{\tabcolsep}{1.5pt} % adjust horizontal space
  \renewcommand{\arraystretch}{1.2} % adjust vertical space
  \newcommand{\quantTit}[1]{\multicolumn{3}{c}{\scriptsize #1}}
    \newcommand{\quantSec}[1]{\scriptsize #1}
    \newcommand{\quantInd}[1]{\scriptsize #1}
    \newcommand{\quantVal}[1]{\scalebox{0.83}[1.0]{$ #1 $}}
    \newcommand{\quantBes}[1]{\scalebox{0.83}[1.0]{$\uline{ #1 }$}}
    
  \centering
  \begin{tabular}{lccccccccccccccccccccc}
    \toprule

    \multirow{4}{*}[-0.28em]{Method}   
     &
     & \multicolumn{4}{c}{Vimeo90K~\cite{xue2019video}}
     & \multicolumn{4}{c}{Middlebury~\cite{baker2011database}}
      & \multicolumn{4}{c}{Vid4~\cite{liu2013bayesian}}
      & \multicolumn{4}{c}{REDS~\cite{Nah_2019_CVPR_Workshops_SR}}
        & \multicolumn{4}{c}{HD~\cite{bao2018memc}}
     
     \\
        \cmidrule(lr){3-6}
        \cmidrule(lr){7-10}
        \cmidrule(lr){11-14}
        \cmidrule(lr){15-18}
        \cmidrule(lr){19-22}
     &
     & \multicolumn{2}{c}{ 4 $\times$}
     & \multicolumn{2}{c}{ 8 $\times$} 
 
     & \multicolumn{2}{c}{ 4 $\times$}
     & \multicolumn{2}{c}{ 8 $\times$} 
     
     & \multicolumn{2}{c}{ 4 $\times$}
     & \multicolumn{2}{c}{ 8 $\times$} 
     
     & \multicolumn{2}{c}{ 4 $\times$}
     & \multicolumn{2}{c}{ 8 $\times$} 
     
     & \multicolumn{2}{c}{ 4 $\times$}
     & \multicolumn{2}{c}{ 8 $\times$} 

     \\
    
        \cmidrule(lr){3-4}
        \cmidrule(lr){5-6}
        \cmidrule(lr){7-8}
        \cmidrule(lr){9-10}
        \cmidrule(lr){11-12}
        \cmidrule(lr){13-14}
        \cmidrule(lr){15-16}
        \cmidrule(lr){17-18}
        \cmidrule(lr){19-20}
        \cmidrule(lr){21-22}
    
    & & \quantSec{PSNR} & \quantSec{SSIM}  & \quantSec{PSNR} & \quantSec{SSIM}  
     & \quantSec{PSNR} & \quantSec{SSIM}  & \quantSec{PSNR} & \quantSec{SSIM} 
     & \quantSec{PSNR} & \quantSec{SSIM}  & \quantSec{PSNR} & \quantSec{SSIM} 
     & \quantSec{PSNR} & \quantSec{SSIM}  & \quantSec{PSNR} & \quantSec{SSIM} 
     & \quantSec{PSNR} & \quantSec{SSIM}  & \quantSec{PSNR} & \quantSec{SSIM} 
    \\
    
    \midrule
        Bicubic 
              & \multirow{2}{*}{\quantSec{1~LR}}
              
              & \quantVal{29.77} &  \quantVal{0.903}
              & \quantVal{25.79} &  \quantVal{0.824}
              
              & \quantVal{27.71}  & \quantVal{0.858}
              & \quantVal{23.82}  & \quantVal{0.763}
              
              & \quantVal{22.37} & \quantVal{0.610}
              & \quantVal{19.82} & \quantVal{0.615}
              
              & \quantVal{26.29} & \quantVal{0.801}
              & \quantVal{23.48} & \quantVal{0.702}
              
              & \quantVal{32.98} & \quantVal{0.936}
              & \quantVal{28.06} & \quantVal{0.880}
              
              \\
        EDSR~\cite{Lim_2017_CVPR_Workshops}
              & 
              & \quantVal{33.11}  & \quantVal{0.941} 
              
              & \quantVal{28.20}  & \quantVal{0.870}
              
              & \quantVal{30.92} & \quantVal{0.905}
              & \quantVal{26.55} & \quantVal{0.813}
              
              & \quantVal{24.06} & \quantVal{0.818}
              & \quantVal{20.80}& \quantVal{0.678}

              & \quantVal{28.51} & \quantVal{0.857}
              & \quantVal{25.05} & \quantVal{0.754}

              & \quantVal{35.68}  & \quantVal{0.957}
              & \quantVal{30.61}  &\quantVal{0.909}
        
              \\
    \midrule          
              
      ToFlow-SR~\cite{xue2019video}
              & \multirow{3}{*}{\quantSec{7~LR}}
              & \quantVal{33.08} & \quantVal{0.942}
              & \quantVal{28.35} & \quantVal{0.882}
              
              & \quantVal{30.27} & \quantVal{0.897}
              & \quantVal{26.02} & \quantVal{0.801}
              
              & \quantVal{24.41} & \quantVal{0.743}
              & \quantVal{21.49} & \quantVal{0.712}
              
              & \quantVal{27.98}  & \quantVal{0.799}
              &  \quantVal{23.59} & \quantVal{0.725}
              
              & \quantVal{33.55} & \quantVal{0.929}
              & \quantVal{27.01} & \quantVal{0.856}
              
              \\
              
         DUF~\cite{jo2018deep}
              &
              & \quantVal{34.33} & \quantVal{0.922}
              & \quantVal{28.55} & \quantVal{0.901}

              & \quantVal{30.89} & \quantVal{0.915}
              & \quantVal{27.46} & \quantVal{0.828}
              
              & \quantVal{25.79} & \quantVal{0.814}
              & \quantVal{21.42} & \quantVal{0.714}  
              
              & \quantVal{28.63} & \quantVal{0.825}
              & \quantVal{24.57} & \quantVal{0.788}
              
              & \quantVal{34.15} & \quantVal{0.926}
              & \quantVal{28.45} & \quantVal{0.891}
              
              \\

        EDVR~\cite{wang2019edvr}
              &
              & \quantVal{35.79} & \quantVal{0.937}
              & \quantVal{29.93} & \quantVal{0.896}
              
              & \quantVal{31.91} & \quantVal{0.924}
              & \quantVal{27.53} & \quantVal{0.832}
              
              & \quantVal{25.83} & \quantVal{0.808}
              & \quantVal{21.49} & \quantVal{0.712}
              
              & \quantVal{31.09} & \quantVal{0.880}
              & \quantVal{25.07} & \quantVal{0.755}  
              
              & \quantVal{34.04} & \quantVal{0.921}
              & \quantVal{29.66} & \quantVal{0.877}
     
              \\

      \midrule
      
      AWnet~\cite{cheng2020dual}
              & \multirow{2}{*}{\quantSec{1 HR~/~1 LR}}
              & \quantVal{39.88}  & \quantVal{\second{0.986}}
              & \quantVal{\second{36.63}}  & \quantVal{\second{0.977}}
              & \quantVal{\second{35.51}}  & \quantVal{\second{0.967}}
              & \quantVal{31.31}  & \quantVal{0.922}
              & \quantVal{\second{30.79}}  & \quantVal{0.943}
              & \quantVal{\second{27.32}}  & \quantVal{\second{0.858}}
              
              & \quantVal{\second{32.67}} & \quantVal{\second{0.937}}
              & \quantVal{31.08} & \quantVal{0.914}
              
              & \quantVal{37.13} & \quantVal{0.969} 
              & \quantVal{32.51} & \quantVal{0.937}

              \\ 
        
    CrossNet~\cite{zheng2018crossnet}
              & 
              & \quantVal{\second{40.85}}  & \quantVal{0.968}
              & \quantVal{36.15}  & \quantVal{\second{0.977}}
              & \quantVal{34.73}  & \quantVal{0.961}
              & \quantVal{\second{32.52}}  & \quantVal{\second{0.932}}
               
              & \quantVal{30.46} & \quantVal{\second{0.958}}
              & \quantVal{27.25} & \quantVal{0.856}
              
              & \quantVal{30.48} & \quantVal{0.869}
              & \quantVal{\second{31.58}} & \quantVal{\second{0.915}}
              
              & \quantVal{\second{37.42}} & \quantVal{\second{0.972}}
              & \quantVal{\second{32.98}} & \quantVal{\second{0.948}}
              
              \\

        \midrule
        PASS-Net~(Ours)
              & \multirow{1}{*}{\quantSec{3 HR~/~1 LR}}
              
              & \quantVal{\first{41.72}} & \quantVal{\first{0.989}}
              & \quantVal{\first{38.81}} & \quantVal{\first{0.982}}
              
              & \quantVal{\first{36.40}} & \quantVal{\first{0.967}}
              & \quantVal{\first{33.66}} & \quantVal{\first{0.948}}
              
              & \quantVal{\first{31.21}} & \quantVal{\first{0.951}}
              & \quantVal{\first{29.44}}  & \quantVal{\first{0.931}}
              
              & \quantVal{\first{34.33}} & \quantVal{\first{0.945}}
              & \quantVal{\first{32.27}} & \quantVal{\first{0.923}}
              
              & \quantVal{\first{38.18}} & \quantVal{\first{0.981}}
              & \quantVal{\first{34.33}} & \quantVal{\first{0.956}}
              
              \\
              
        \bottomrule

  \end{tabular}
    % \vspace{-5pt}  
\end{table*}

    \subsection{Comparisons with the State-of-the-arts}
    
    We evaluate the proposed PASS-Net against the following state-of-the-arts algorithms, including BeyondMSE~\cite{mathieu2015deep},  DVF~\cite{liu2017video},  CtrlGen~\cite{hao2018controllable} for video prediction and~EDSR~\cite{Lim_2017_CVPR_Workshops}, ToFlow-SR~\cite{xue2019video}, DUF~\cite{jo2018deep}, EDVR~\cite{wang2019edvr}, AWnet~\cite{cheng2020dual}, CrossNet~\cite{zheng2018crossnet} for video supre-resolution.
    We compare our model with the state-of-the-art algorithms using the regular $\times 4$ and $\times 8$ settings of video super-resolution.
    For video prediction, we test the model using full-resolution frames.
    
    \Paragraph{Video Frame Prediction.}
    In~\tabref{tab:compare_prediction}, we show comparisons on the UCF101~\cite{soomro2012ucf101} and Vimeo90K~\cite{xue2019video} dataset for the task of next frame prediction.
    Our algorithm performs favorably against existing methods in the compared dataset, especially on the Vimeo90K~\cite{xue2019video} dataset with a $2.75$dB gain over DVF~\cite{liu2017video} in terms of PSNR.

    \Paragraph{Video Frame Super-Resolution.}
    As shown in~\tabref{tab:compare_next_frame_sr}, we compare our model with multiple kinds of frame super-resolution algorithms.
    Our model outperforms all other methods for all evaluation datasets in terms of PSNR and SSIM metrics.
    Our model performs favorably against the non-reference video super-resolution methods (\eg EDVR~\cite{wang2019edvr}) mainly because our model transfers high-resolution textures from the given reference images to produce outputs with rich textual details.
    Our approach outperforms the reference-based video super-resolution methods (\eg AWnet~\cite{cheng2020dual}, CrossNet~\cite{zheng2018crossnet}) mainly because we adopt the flow prediction process to assistant the frame fusion module to predict an accurate flow, thus the frame synthesize module could produce high-quality frames.
     
    Moreover, we test the model using frames with multiple gaps, which is more piratical in real applications. 
    For example, there is a 7-frames group of each clip in the Vimeo90K dataset.
    In the previous section, we train our model using the first three frames as high-resolution reference frames, and the down-scaled fourth frame as the low-resolution current input.
    In this case, the frame gap is equal to one.
    In this section, we test our model for multiple frame gaps, namely, we use the first three high-resolution frames as a reference and the $n\text{th}$ down-scaled frame as the current input.

    \begin{table}[t]
  \caption{
     \textbf{Quantitative comparisons on the next frame prediction on the UCF101 and Vimeo-90K test set.}
  }
%   \vspace{-1mm}
  \label{tab:compare_prediction}
  \footnotesize
  \renewcommand{\tabcolsep}{10pt} % adjust horizontal space
  \renewcommand{\arraystretch}{1.2} % adjust vertical space
\newcommand{\quantTit}[1]{\multicolumn{3}{c}{\scriptsize #1}}
    \newcommand{\quantSec}[1]{\scriptsize #1}
    \newcommand{\quantInd}[1]{\scriptsize #1}
    \newcommand{\quantVal}[1]{\scalebox{0.83}[1.0]{$ #1 $}}
    \newcommand{\quantBes}[1]{\scalebox{0.83}[1.0]{$\uline{ #1 }$}}
  \centering
  \begin{tabular}{lcccc}
    \toprule
    
    \multirow{2}{*}[-0.28em]{Method}   
    %  & Runtime 
    % & Parameters 
     &  \multicolumn{2}{c}{UCF101~\cite{soomro2012ucf101}}
      & \multicolumn{2}{c}{Vimeo-90K~\cite{xue2019video}} \\
    
    %  \cmidrule(l{2pt}r{1pt}){2-3}
    % \cmidrule(l{2pt}r{1pt}){4-5} 
    \cmidrule(lr){2-3}
    \cmidrule(lr){4-5}
    
    % % %\cmidrule(r){2-3} \cmidrule(r){4-5} \cmidrule(r){a-b}
    & \quantSec{PSNR} & \quantSec{SSIM}  & 
    \quantSec{PSNR} & \quantSec{SSIM} \\
    \midrule
        BeyondMSE~\cite{mathieu2015deep}
              & \quantVal{30.01}  & \quantVal{0.897}
              & \quantVal{26.94}  & \quantVal{0.850}         \\
              
        DVF~\cite{liu2017video}
              & \quantVal{30.29}  & \quantVal{0.901}
              & \quantVal{27.55}  & \quantVal{0.866}    \\ % 
              
        CtrlGen~\cite{hao2018controllable}
              & \quantVal{28.13}   & \quantVal{0.864}
              & \quantVal{26.32}   & \quantVal{0.832}  \\
        
        % \midrule
        
        PASS-Net~-~w/o prop
              & \quantVal{\second{30.50}}  &   \quantVal{\second{0.901}}
              & \quantVal{\second{29.71}}    & \quantVal{\second{0.887}}  \\
        PASS-Net (Ours)
              & \quantVal{\first{31.05}}    & \quantVal{\first{0.910}}
              & \quantVal{\first{30.30}}    & \quantVal{\first{0.899}}  \\

        \bottomrule

  \end{tabular}
    \vspace{-10pt}  
\end{table}

\begin{figure}[t]
	\footnotesize
    % \scriptsize

	\centering
% 	\renewcommand{\tabcolsep}{1.0pt} % adjust horizontal space
% 	\renewcommand{\arraystretch}{1.0} % adjust vertical space
% 	\begin{tabular}{c}
	\includegraphics[width= 0.47\textwidth]{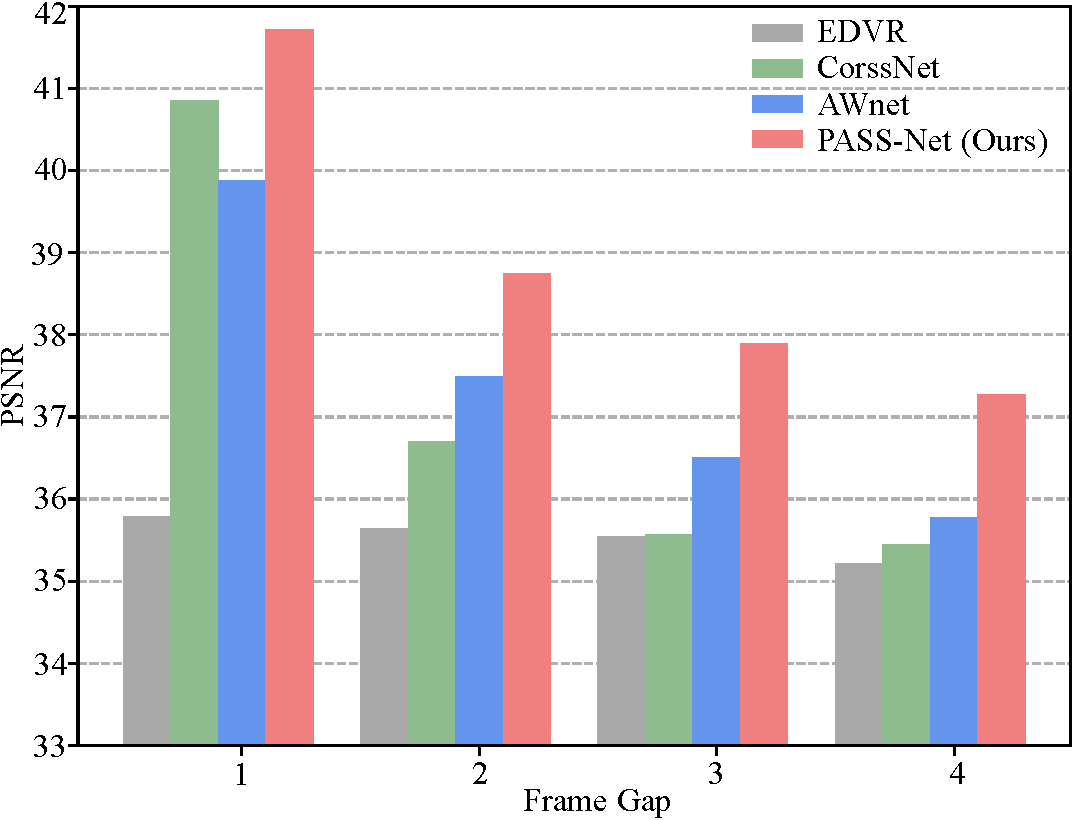}
	
% \end{tabular}
 	\vspace{-5pt}
	\caption{
		\textbf{Comparisons of video super-resolution using different frame gaps on the Vimeo90K evaluation dataset~\cite{xue2019video}.}
		Our model is trained using a frame gap equals to one, and we test the model using frames with multiple gaps.
	}
 	\vspace{-10pt}
	\label{fig:multiple_gap} 
\end{figure}

    As shown in~\figref{fig:multiple_gap}, our model performs favorably against all the compared methods.
    Even trained with frame gap equals to one, the reference-based models (\eg CrossNet, AWnet, and our algorithm) always outperform non-reference based methods~(\eg EDVR).
    However, with the increase of the frame gap, the PSNR of the synthesized frame also decreases, because the long temporal distance enlarges the objection motion, resulting in inaccurate motion estimation.
    In the meantime, the content similarity of the historical reference frames and the current frame is also reduced, making it hard to synthesize high-quality frames.
    Moreover, we believe that if we train the model using a frame with a randomly selected frame gap, our model will be more robust to multi-gap frame synthesis.

\section{Conclusion}
    In this work, we propose to enhance video quality using lossy frames in two streaming patterns, including the lost case, and the lossy case.
    For the lost case,  we propose to use previously received high-resolution images to predict future frames.
    For the lossy case, we propose to use previously received high-resolution frames to enhance the low-quality current frames.
    Our proposed video enhancement model has a unified framework both for video prediction and video super-resolution.
    Extensive experimental results demonstrate that our method performs favorably against state-of-the-art methods in the lossy video streaming environment.
    However, our model can not produce images in real-time, we will focus on speeding up the algorithm in future work.

\clearpage

{\small
\bibliographystyle{ieee_fullname}
\bibliography{mybib}

\begin{thebibliography}{10}\itemsep=-1pt

\bibitem{baker2011database}
Simon Baker, Daniel Scharstein, JP Lewis, Stefan Roth, Michael~J Black, and
  Richard Szeliski.
\newblock A database and evaluation methodology for optical flow.
\newblock {\em IJCV}, 2011.

\bibitem{balachandran2013developing}
Athula Balachandran, Vyas Sekar, Aditya Akella, Srinivasan Seshan, Ion Stoica,
  and Hui Zhang.
\newblock Developing a predictive model of quality of experience for internet
  video.
\newblock {\em ACM SIGCOMM Computer Communication Review}, 2013.

\bibitem{bao2019depth}
Wenbo Bao, Wei-Sheng Lai, Chao Ma, Xiaoyun Zhang, Zhiyong Gao, and Ming-Hsuan
  Yang.
\newblock Depth-aware video frame interpolation.
\newblock In {\em CVPR}, 2019.

\bibitem{bao2018memc}
Wenbo Bao, Wei-Sheng Lai, Xiaoyun Zhang, Zhiyong Gao, and Ming-Hsuan Yang.
\newblock Memc-net: Motion estimation and motion compensation driven neural
  network for video interpolation and enhancement.
\newblock {\em TPAMI}, 2019.

\bibitem{sintel}
D.~J. Butler, J. Wulff, G.~B. Stanley, and M.~J. Black.
\newblock A naturalistic open source movie for optical flow evaluation.
\newblock In {\em ECCV}, 2012.

\bibitem{byeon2018contextvp}
Wonmin Byeon, Qin Wang, Rupesh Kumar~Srivastava, and Petros Koumoutsakos.
\newblock Contextvp: Fully context-aware video prediction.
\newblock In {\em ECCV}, 2018.

\bibitem{caballero2017real}
Jose Caballero, Christian Ledig, Andrew Aitken, Alejandro Acosta, Johannes
  Totz, Zehan Wang, and Wenzhe Shi.
\newblock Real-time video super-resolution with spatio-temporal networks and
  motion compensation.
\newblock In {\em CVPR}, 2017.

\bibitem{charbonnier1994two}
Pierre Charbonnier, Laure Blanc-Feraud, Gilles Aubert, and Michel Barlaud.
\newblock Two deterministic half-quadratic regularization algorithms for
  computed imaging.
\newblock In {\em ICCV}, 1994.

\bibitem{cheng2020dual}
Ming Cheng, Zhan Ma, Salman Asif, Yiling Xu, Haojie Liu, Wenbo Bao, and Jun
  Sun.
\newblock A dual camera system for high spatiotemporal resolution video
  acquisition.
\newblock {\em TPAMI}, 2020.

\bibitem{denton2018stochastic}
Emily Denton and Rob Fergus.
\newblock Stochastic video generation with a learned prior.
\newblock In {\em ECCV}, 2018.

\bibitem{dobrian2011understanding}
Florin Dobrian, Vyas Sekar, Asad Awan, Ion Stoica, Dilip Joseph, Aditya Ganjam,
  Jibin Zhan, and Hui Zhang.
\newblock Understanding the impact of video quality on user engagement.
\newblock {\em ACM SIGCOMM}, 2011.

\bibitem{dong2015image}
Chao Dong, Chen~Change Loy, Kaiming He, and Xiaoou Tang.
\newblock Image super-resolution using deep convolutional networks.
\newblock {\em TPAMI}, 2015.

\bibitem{dong2016accelerating}
Chao Dong, Chen~Change Loy, and Xiaoou Tang.
\newblock Accelerating the super-resolution convolutional neural network.
\newblock In {\em ECCV}, 2016.

\bibitem{gao2019disentangling}
Hang Gao, Huazhe Xu, Qi-Zhi Cai, Ruth Wang, Fisher Yu, and Trevor Darrell.
\newblock Disentangling propagation and generation for video prediction.
\newblock In {\em CVPR}, 2019.

\bibitem{hao2018controllable}
Zekun Hao, Xun Huang, and Serge Belongie.
\newblock Controllable video generation with sparse trajectories.
\newblock In {\em CVPR}, 2018.

\bibitem{he2016deep}
Kaiming He, Xiangyu Zhang, Shaoqing Ren, and Jian Sun.
\newblock Deep residual learning for image recognition.
\newblock In {\em CVPR}, 2016.

\bibitem{huang2017densely}
Gao Huang, Zhuang Liu, Laurens Van Der~Maaten, and Kilian~Q Weinberger.
\newblock Densely connected convolutional networks.
\newblock In {\em CVPR}, 2017.

\bibitem{IMKDB17}
Eddy Ilg, Nikolaus Mayer, Tonmoy Saikia, Margret Keuper, Alexey Dosovitskiy,
  and Thomas Brox.
\newblock Flownet 2.0: Evolution of optical flow estimation with deep networks.
\newblock In {\em CVPR}, 2017.

\bibitem{index2016white}
Cisco Visual~Networking Index.
\newblock White paper: Cisco vni forecast and methodology, 2015-2020.
\newblock
  \url{http://www.webvideomarketing.org/pdf/Cisco_Video_and_Visual_Networking_Index_Report_8.10.16.pdf},
  2016.

\bibitem{jiang2018super}
Huaizu Jiang, Deqing Sun, Varun Jampani, Ming-Hsuan Yang, Erik Learned-Miller,
  and Jan Kautz.
\newblock Super slomo: High quality estimation of multiple intermediate frames
  for video interpolation.
\newblock In {\em CVPR}, 2018.

\bibitem{jo2018deep}
Younghyun Jo, Seoung Wug~Oh, Jaeyeon Kang, and Seon Joo~Kim.
\newblock Deep video super-resolution network using dynamic upsampling filters
  without explicit motion compensation.
\newblock In {\em CVPR}, 2018.

\bibitem{kappeler2016video}
Armin Kappeler, Seunghwan Yoo, Qiqin Dai, and Aggelos~K Katsaggelos.
\newblock Video super-resolution with convolutional neural networks.
\newblock {\em IEEE Transactions on Computational Imaging}, 2016.

\bibitem{kim2016accurate}
Jiwon Kim, Jung Kwon~Lee, and Kyoung Mu~Lee.
\newblock Accurate image super-resolution using very deep convolutional
  networks.
\newblock In {\em CVPR}, 2016.

\bibitem{kim2016deeply}
Jiwon Kim, Jung Kwon~Lee, and Kyoung Mu~Lee.
\newblock Deeply-recursive convolutional network for image super-resolution.
\newblock In {\em CVPR}, 2016.

\bibitem{kim2018spatio}
Tae~Hyun Kim, Mehdi~SM Sajjadi, Michael Hirsch, and Bernhard Sch{\"o}lkopf.
\newblock Spatio-temporal transformer network for video restoration.
\newblock In {\em ECCV}, 2018.

\bibitem{kingma2014adam}
Diederik~P Kingma and Jimmy Ba.
\newblock Adam: A method for stochastic optimization.
\newblock {\em arXiv}, 2014.

\bibitem{krishnan2013video}
S~Shunmuga Krishnan and Ramesh~K Sitaraman.
\newblock Video stream quality impacts viewer behavior: inferring causality
  using quasi-experimental designs.
\newblock {\em TON}, 2013.

\bibitem{ledig2017photo}
Christian Ledig, Lucas Theis, Ferenc Husz{\'a}r, Jose Caballero, Andrew
  Cunningham, Alejandro Acosta, Andrew Aitken, Alykhan Tejani, Johannes Totz,
  Zehan Wang, et~al.
\newblock Photo-realistic single image super-resolution using a generative
  adversarial network.
\newblock In {\em CVPR}, 2017.

\bibitem{liao2015video}
Renjie Liao, Xin Tao, Ruiyu Li, Ziyang Ma, and Jiaya Jia.
\newblock Video super-resolution via deep draft-ensemble learning.
\newblock In {\em ICCV}, 2015.

\bibitem{Lim_2017_CVPR_Workshops}
Bee Lim, Sanghyun Son, Heewon Kim, Seungjun Nah, and Kyoung Mu~Lee.
\newblock Enhanced deep residual networks for single image super-resolution.
\newblock In {\em CVPR Workshops}, 2017.

\bibitem{liu2013bayesian}
Ce Liu and Deqing Sun.
\newblock On bayesian adaptive video super resolution.
\newblock {\em TPAMI}, 2013.

\bibitem{liu2017robust}
Ding Liu, Zhaowen Wang, Yuchen Fan, Xianming Liu, Zhangyang Wang, Shiyu Chang,
  and Thomas Huang.
\newblock Robust video super-resolution with learned temporal dynamics.
\newblock In {\em ICCV}, 2017.

\bibitem{liu2018non}
Ding Liu, Bihan Wen, Yuchen Fan, Chen~Change Loy, and Thomas~S Huang.
\newblock Non-local recurrent network for image restoration.
\newblock In {\em NIPS}, 2018.

\bibitem{liu2018future}
Wen Liu, Weixin Luo, Dongze Lian, and Shenghua Gao.
\newblock Future frame prediction for anomaly detection--a new baseline.
\newblock In {\em CVPR}, 2018.

\bibitem{liu2017video}
Ziwei Liu, Raymond~A Yeh, Xiaoou Tang, Yiming Liu, and Aseem Agarwala.
\newblock Video frame synthesis using deep voxel flow.
\newblock In {\em CVPR}, 2017.

\bibitem{lotter2016deep}
William Lotter, Gabriel Kreiman, and David Cox.
\newblock Deep predictive coding networks for video prediction and unsupervised
  learning.
\newblock In {\em ICLR}, 2017.

\bibitem{mao2017neural}
Hongzi Mao, Ravi Netravali, and Mohammad Alizadeh.
\newblock Neural adaptive video streaming with pensieve.
\newblock In {\em ACM SIGCOMM}, 2017.

\bibitem{mathieu2015deep}
Michael Mathieu, Camille Couprie, and Yann LeCun.
\newblock Deep multi-scale video prediction beyond mean square error.
\newblock In {\em ICLR}, 2016.

\bibitem{Nah_2019_CVPR_Workshops_SR}
Seungjun Nah, Radu Timofte, Shuhang Gu, Sungyong Baik, Seokil Hong, Gyeongsik
  Moon, Sanghyun Son, and Kyoung~Mu Lee.
\newblock Ntire 2019 challenge on video super-resolution: Methods and results.
\newblock In {\em CVPR Workshops}, June 2019.

\bibitem{niklaus2018context}
Simon Niklaus and Feng Liu.
\newblock Context-aware synthesis for video frame interpolation.
\newblock In {\em CVPR}, 2018.

\bibitem{paliwal2020deep}
Avinash Paliwal and Nima~Khademi Kalantari.
\newblock Deep slow motion video reconstruction with hybrid imaging system.
\newblock {\em TPAMI}, 2020.

\bibitem{reda2018sdc}
Fitsum~A Reda, Guilin Liu, Kevin~J Shih, Robert Kirby, Jon Barker, David
  Tarjan, Andrew Tao, and Bryan Catanzaro.
\newblock Sdc-net: Video prediction using spatially-displaced convolution.
\newblock In {\em ECCV}, 2018.

\bibitem{ren2019fusion}
Zhile Ren, Orazio Gallo, Deqing Sun, Ming-Hsuan Yang, Erik Sudderth, and Jan
  Kautz.
\newblock A fusion approach for multi-frame optical flow estimation.
\newblock In {\em WACV}, 2019.

\bibitem{sajjadi2018frame}
Mehdi~SM Sajjadi, Raviteja Vemulapalli, and Matthew Brown.
\newblock Frame-recurrent video super-resolution.
\newblock In {\em CVPR}, 2018.

\bibitem{schwarz2007overview}
Heiko Schwarz, Detlev Marpe, and Thomas Wiegand.
\newblock Overview of the scalable video coding extension of the h. 264/avc
  standard.
\newblock {\em TCSVT}, 2007.

\bibitem{soomro2012ucf101}
Khurram Soomro, Amir~Roshan Zamir, and Mubarak Shah.
\newblock Ucf101: A dataset of 101 human actions classes from videos in the
  wild.
\newblock {\em arXiv}, 2012.

\bibitem{sun2018pwc}
Deqing Sun, Xiaodong Yang, Ming-Yu Liu, and Jan Kautz.
\newblock Pwc-net: Cnns for optical flow using pyramid, warping, and cost
  volume.
\newblock In {\em CVPR}, 2018.

\bibitem{tian2012towards}
Guibin Tian and Yong Liu.
\newblock Towards agile and smooth video adaptation in dynamic http streaming.
\newblock In {\em ACM SIGCOMM}, 2012.

\bibitem{tian2020tdan}
Yapeng Tian, Yulun Zhang, Yun Fu, and Chenliang Xu.
\newblock Tdan: Temporally-deformable alignment network for video
  super-resolution.
\newblock In {\em CVPR}, 2020.

\bibitem{tong2017image}
Tong Tong, Gen Li, Xiejie Liu, and Qinquan Gao.
\newblock Image super-resolution using dense skip connections.
\newblock In {\em ICCV}, 2017.

\bibitem{wang2019edvr}
Xintao Wang, Kelvin~CK Chan, Ke Yu, Chao Dong, and Chen Change~Loy.
\newblock Edvr: Video restoration with enhanced deformable convolutional
  networks.
\newblock In {\em CVPR}, 2019.

\bibitem{xu2019quadratic}
Xiangyu Xu, Li Siyao, Wenxiu Sun, Qian Yin, and Ming-Hsuan Yang.
\newblock Quadratic video interpolation.
\newblock In {\em NIPS}, 2019.

\bibitem{xue2019video}
Tianfan Xue, Baian Chen, Jiajun Wu, Donglai Wei, and William~T Freeman.
\newblock Video enhancement with task-oriented flow.
\newblock {\em IJCV}, 2019.

\bibitem{yang2020learning}
Fuzhi Yang, Huan Yang, Jianlong Fu, Hongtao Lu, and Baining Guo.
\newblock Learning texture transformer network for image super-resolution.
\newblock In {\em CVPR}, 2020.

\bibitem{yin2015control}
Xiaoqi Yin, Abhishek Jindal, Vyas Sekar, and Bruno Sinopoli.
\newblock A control-theoretic approach for dynamic adaptive video streaming
  over http.
\newblock In {\em ACM SIGCOMM}, 2015.

\bibitem{zhang2018image}
Yulun Zhang, Kunpeng Li, Kai Li, Lichen Wang, Bineng Zhong, and Yun Fu.
\newblock Image super-resolution using very deep residual channel attention
  networks.
\newblock In {\em ECCV}, 2018.

\bibitem{zhang2018residual}
Yulun Zhang, Yapeng Tian, Yu Kong, Bineng Zhong, and Yun Fu.
\newblock Residual dense network for image super-resolution.
\newblock In {\em CVPR}, 2018.

\bibitem{zheng2018crossnet}
Haitian Zheng, Mengqi Ji, Haoqian Wang, Yebin Liu, and Lu Fang.
\newblock Crossnet: An end-to-end reference-based super resolution network
  using cross-scale warping.
\newblock In {\em ECCV}, 2018.

\end{thebibliography}
}

\end{document}

%%%%%%%%% TITLE
%\title{Prediction-assistant Video Frame Super-Resolution}

\title{Prediction-assistant Frame Super-Resolution for Video Streaming}

% \and

\author{Wang Shen$^1$ 
\hspace{3pt}
Wenbo Bao$^1$
\hspace{3pt}
Guangtao Zhai$^1$ $^\ast$
\hspace{3pt}
Charlie L Wang$^2$
\hspace{3pt} 
Jerry W Hu$^2$
\hspace{3pt} 
Zhiyong Gao$^1$
\\
$^1$ Institute of Image Communication and Network Engineering, \\ Shanghai Jiao Tong University \\
$^2$ Intel
\vspace{-6mm}
}

\maketitle

\section{Flow Propagation}

   \begin{figure*}[h]
	\footnotesize
    % \scriptsize

	\centering
% 	\renewcommand{\tabcolsep}{1.0pt} % adjust horizontal space
% 	\renewcommand{\arraystretch}{1.0} % adjust vertical space
% 	\begin{tabular}{c}
	\includegraphics[width= 0.6\textwidth]{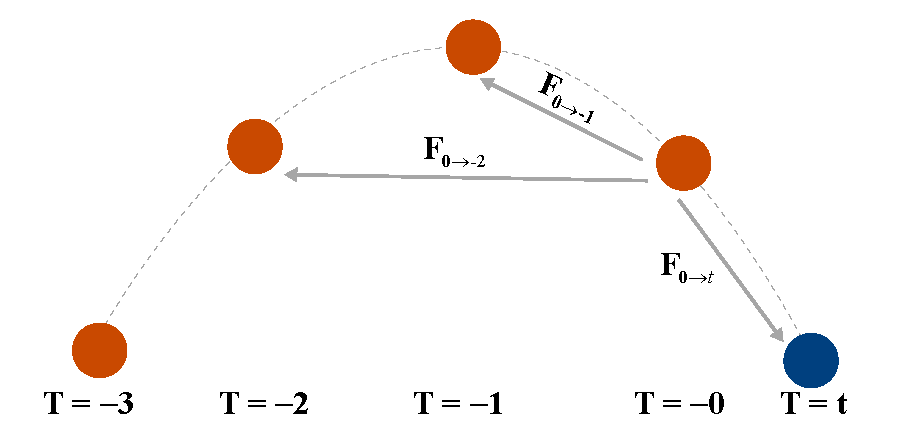}
	
% \end{tabular}
%  	\vspace{-5pt}
	\caption{
		\textbf{Illustration of flow propagation.} 
	}
%  	\vspace{-10pt}
	\label{fig:flow_propa} 
\end{figure*}

    In this supplementary document, we introduce the process of flow propagation.
    Similar to the process of flow interpolation~\cite{xu2019quadratic}, the flow propagation is shown in~\figref{fig:flow_propa}.
    This figure shows a pixel is doing projectile motion (i.e., non-linear motion).
    We are given the position of pixels at $T=\{-2,-1,0\}$, we want to compute the position of the pixel at time $t$.
    Under the assumption of the uniform acceleration~\cite{xu2019quadratic}, these optical flow fields can be modeled using $\mathbf{F}_{0 \rightarrow m} =\mathbf{v}_0 \cdot m + \frac{1}{2}\cdot \mathbf{a} \cdot m^2$, where $m \in [-1, -2]$, $\mathbf{v}_0$ and $\mathbf{a}$ denote speed and acceleration respectively.
    \begin{align}
        \mathbf{F}_{0 \rightarrow -1}  &= \mathbf{v}_0 \cdot -1 + \frac{1}{2} \cdot \mathbf{a} \cdot 1^2, 
        \\
        \mathbf{F}_{0 \rightarrow -2}  &= \mathbf{v}_0 \cdot -2 + \frac{1}{2} \cdot \mathbf{a} \cdot 2^2.
    \end{align}
    The predicted flow at time $t$ can also be formulated as:
    \begin{equation}
        \mathbf{F}_{0 \rightarrow t}  = \mathbf{v}_0 \cdot t + \frac{1}{2} \cdot \mathbf{a} \cdot t^2.
    \end{equation}
    Then we eliminate $\mathbf{v}_0$ and $\mathbf{a}$, we get the propagated flow:
     \begin{equation}
        \label{eq:flow2}
        \mathbf{\tilde{F}}_{0 \rightarrow t} = 0.5t(t+1)\mathbf{F}_{0 \rightarrow -2} - t(t+2)\mathbf{F}_{0 \rightarrow -1}.
    \end{equation}
    
% \clearpage

\vspace{25pt}

{\small
\bibliographystyle{ieee_fullname}
\bibliography{mybib}
}